# UBMF: Uncertainty-Aware Bayesian Meta-Learning Framework for Fault Diagnosis with Imbalanced Industrial Data

Zhixuan Lian[1]; Shangyu Li[1,2,3,4]; Qixuan Huang[2,3,4]; Zijian Huang[2,3,4]; Haifei Liu[2,3,4]; Jianan Qiu[1]; Puyu Yang[2,3,4]; Laifa Tao [1,2,3,4]

[1] Hangzhou International Innovation Institute, Beihang University, China
[2] Institute of Reliability Engineering, Beihang University, Beijing, China
[3] Science & Technology on Reliability & Environmental Engineering Laboratory, Beijing, China
[4] School of Reliability and Systems Engineering, Beihang University, Beijing, China
*Corresponding author, taolaifa@buaa.edu.cn

Zhixuan Lian:Methodology, Software Data curation, Writing- Original draft preparation. Shangy Li: Visualization, Investigation. Qixuan Huang:Supervision.:Software, Validation.: Zijian Huang:Supervision.:Software, Validation. Haifei Liu:Writing- Reviewing and Editing, Jianan Qiu:Writing- Reviewing and Editing,Puyu Yang:Writing- Reviewing and Editing; Laifa Tao:Conceptualization,

**Abstract**

Fault diagnosis of mechanical equipment involves data collection, feature extraction, and pattern recognition but is often hindered by the imbalanced nature of industrial data, introducing significant uncertainty and reducing diagnostic reliability. To address these challenges, this study proposes the Uncertainty-Aware Bayesian Meta-Learning Framework (UBMF), which integrates four key modules: data perturbation injection for enhancing feature robustness, cross-task self-supervised feature extraction for improving transferability, uncertainty-based sample filtering for robust out-of-domain generalization, and Bayesian meta-knowledge integration for fine-grained classification. Experimental results on ten open-source datasets under various imbalanced conditions, including cross-task, small-sample, and unseen-sample scenarios, demonstrate the superiority of UBMF, achieving an average improvement of 42.22% across ten Any-way 1-5-shot diagnostic tasks. This integrated framework effectively enhances diagnostic accuracy, generalization, and adaptability, providing a reliable solution for complex industrial fault diagnosis.

**KEYWORDS**

Fault Diagnosis; Bayesian Framework; Meta-Learning ; Imbalanced Industrial Data; Uncertainty

## 1. Introduction

The manufacturing industry is the cornerstone of national economies (Yuan et al., 2023), driven by innovation and strategic initiatives such as China's "Made in China (2025)," (Xu et al., 2018) Germany's "National Industrial Strategy 2030 (Germann, 2023)," and the U.S.'s "Advanced Manufacturing Leadership Strategy (Benedict et al., 2022)." Mechanical equipment, a core component widely used in high-tech and transportation sectors, is advancing toward greater precision, efficiency, automation, and complexity. However, growing operational demands and harsh conditions lead to performance degradation and failures, highlighting the critical need for real-time monitoring and reliable fault diagnosis.

Zhixuan Lian: lianzhixuan@buaa.edu.cn;0000-0003-0878-9344
Shangyu Li: lishangyu@buaa.edu.cn; 0000-0003-1426-9206;
Qixuan Huang: qxhuang@buaa.edu.cn; 0000-0002-5376-9927;
Zijian Huang: hzj1234@buaa.edu.cn; 0009-0007-5442-7017;
Haifei Liu: phoebeliu@buaa.edu.cn; 0000-0002-3099-7185;
Jianan Qiu: ZY2457230@buaa.edu.cn; 0009-0009-5784-200X;
Laifa Tao: taolaifa@buaa.edu.cn.;0000-0002-9232-7216
Puyu Yang: 20212202@cqu.edu.cn

Fault diagnosis of mechanical equipment is a complex process involving data collection, feature extraction, and pattern recognition (Bajaj et al., 2023). In practical industrial scenarios, the imbalanced nature of industrial data introduces significant uncertainty into this process, posing a critical challenge for achieving accurate and reliable fault diagnosis (Chen et al., 2022). Equipment faults are inherently rare, resulting in a severe imbalance between normal operating data and fault condition data. Imbalanced data amplifies uncertainty in feature extraction and classification (Liu et al., 2022), as models often struggle to capture the intrinsic structure of minority classes or adapt to rare and novel fault scenarios (Zhang et al., 2023). This limitation significantly impacts diagnostic accuracy, particularly in capturing subtle fault patterns or addressing complex and dynamic conditions. Moreover, while models with strong generalization capabilities can handle diverse scenarios, their diagnostic precision is often constrained by the lack of task-specific tuning, leading to suboptimal fault identification in intricate or unique operational contexts. On the other hand, the highly coupled structures and complex functionalities of mechanical equipment (Wang et al., 2022)—characterized by multi-physical fields and multi-objective constraints—demand tailored fault diagnosis models for high accuracy. These models are often customized for specific equipment types and operating conditions, inherently limiting their generalization across diverse environments (Azari et al., 2025). Additionally, the randomness, diversity, and cross-coupling characteristics of fault patterns arising from varying operating environments, such as extreme temperatures, pressures, and dynamic loads, further challenge generalization. While knowledge-driven approaches offer broader applicability, their reliance on incomplete domain knowledge reduces precision, highlighting the trade-off between generalization and task-specific diagnostic performance (Xiao et al., 2024; Yin et al., 2022).

Balancing precision and generalization in fault diagnosis is inherently challenging (Ren et al., 2022), as achieving high accuracy often requires task-specific customization, limiting adaptability, while broader generalization can compromise precision (Tan et al., 2023). This trade-off, compounded by data imbalance-induced uncertainty, underscores the need for innovative solutions to meet the demands of complex industrial environments. To address these challenges, a promising solution has emerged in recent years: the integration of Bayesian inference frameworks (Chan et al., 2024; Feng et al., 2024; Iaiani et al., 2025; Liu et al., 2025). Bayesian approaches offer a unique advantage in balancing high accuracy and strong generalization by incorporating probabilistic reasoning to model uncertainties, enabling robust adaptation to new scenarios with limited data. This paradigm effectively addresses data-driven limitations by parameterizing models that flexibly handle diverse physical conditions while maintaining precision (Chien et al., 2023). Simultaneously, it leverages meta-knowledge to encode prior domain expertise and enhance feature representations, improving the reliability of diagnostics in complex industrial environments (Liu & Wu, 2024). By dynamically adapting to task-specific needs without sacrificing broad applicability, Bayesian frameworks provide a robust foundation for fault diagnosis that bridges the gap between high accuracy and generalization.

In the Bayesian diagnostic framework, data imbalance (Feng et al., 2024; Liu & Yu, 2023) significantly impacts both inner-layer training and outer-layer generalization. In the inner-layer training, data imbalance exacerbates the uncertainty of sample data, causing feature extraction networks to favor majority-class features while neglecting subtle patterns in minority classes. Additionally, conventional loss functions poorly adapt to imbalanced distributions, resulting in asymmetric learning of features (Wong & Tsai, 2021). Furthermore, data imbalance disrupts the Bayesian inference process, leading to posterior probability estimates biased toward majority classes, reducing sensitivity to minority-class samples, and limiting the effective integration of prior information. In the outer-layer training, data imbalance weakens the model's cross-task generalization ability, where cross-task refers to adapting the model from tasks trained on specific operating conditions (e.g., known fault modes or environments) to new, unseen conditions (e.g., unknown fault modes or operational settings) (Song et al., 2025; Wang et al., 2021). This imbalance causes features to become overly dependent on class-specific patterns, hindering the model's ability to generalize across different fault modes or operating scenarios. Simultaneously, the model exhibits insufficient robustness to out-of-distribution (OOD) samples (Fang et al., 2024; Lin et al., 2022), often confusing known and unknown categories, with reduced confidence evaluation and rejection capabilities. To address these issues, A framework (UBMF in short) is proposed, introduces meta-knowledge to guide the diagnostic process, incorporating uncertainty quantification (Li et al., 2025) to improve sample filtering and modeling strategies. By leveraging high-quality samples and correcting feature biases, the framework enhances diagnostic precision, robustness, and generalization in complex industrial environments.

The systematic optimization required to address these challenges involves the following four key modules:

**1. Data Perturbation Injection Based on Prior Knowledge** (Zhang et al., 2024)**:** Addressing inner-layer training, this module enhances feature diversity and mitigates class-specific biases by injecting perturbations informed by prior knowledge, improving feature extraction robustness under imbalanced data conditions.

**2. Cross-task Self-supervised Feature Extraction Enhancement:** Focused on outer-layer training, this module leverages self-supervised learning to strengthen cross-task feature representations, enabling robust parameter optimization and improved transferability across diverse operational conditions.

**3. Sample Filtering Mechanism Based on Uncertainty Metrics:** Targeting out-of-domain (OOD) generalization, this module employs uncertainty-based filtering to remove unreliable samples, enhancing the model's capacity for accurate OOD detection and robust diagnostic reliability in unseen domains.

**4. Classifier Enhancement Based on Bayesian Meta-Knowledge Extraction：** Concentrating on inner-layer training, this module improves fine-grained fault classification by integrating Bayesian meta-knowledge, refining posterior probability calibration and enhancing the classifier's precision under imbalanced scenarios.

By integrating these modules, the framework comprehensively addresses the aforementioned challenges, significantly enhancing the accuracy and generalization of diagnostic models under imbalanced data conditions. The contributions are following:

**1. Introduction of the Uncertainty Modeling Module and Task Balancing and OOD Detection Module** (Cao et al., 2024; Das et al., 2024; Dong et al., 2024; Gital & Bilgen, 2024; Jiang et al., 2025; Lee & Kang, 2024; Lin et al., 2025; Liu et al., 2024)**：** Accurate quantification of predictive uncertainty is critical for PHM decision-making and risk reduction. This work systematically addresses uncertainties arising from unseen conditions, unseen samples, and unseen domains in fault diagnosis by introducing an integrated framework for theoretical modeling, mathematical quantification, and model calibration. Unlike existing approaches that often incorporate uncertainty metrics in a fragmented or task-specific manner, our method provides a cohesive and comprehensive solution, establishing a systematic theoretical foundation for uncertainty management. The proposed uncertainty modeling module unifies the conceptualization of uncertainty sources, employs advanced probabilistic and Bayesian techniques for mathematical modeling and quantification, and ensures calibration to align predictions with true confidence levels. This enables precise detection of unknown anomalies and effective implementation of rejection diagnostics. By incorporating these robust uncertainty metrics, the framework not only enhances the reliability of diagnostics but also introduces an adaptive sample filtering mechanism tailored to the demands of complex industrial scenarios. This integration marks a significant innovation in uncertainty modeling for fault diagnosis, bridging the gaps left by piecemeal approaches and advancing the robustness, adaptability, and decision-making reliability of predictive systems under uncertain conditions.

**2. Development of a Bayesian Framework for Imbalanced Fault Diagnosis：** Using bearings as a case study, this work develops a Bayesian meta-learning framework tailored to imbalanced fault diagnosis scenarios. The framework effectively addresses three types of data imbalance: cross-condition, small-sample, and cross-domain. It significantly enhances diagnostic accuracy in generalized fault diagnosis tasks, providing robust support for complex industrial applications.

**3. Proposal of the Unlabeled Data Pseudo-Labeling Module and Feature Extraction and Generalization Enhancement Module：** These modules improve data balance through pseudo-label propagation, enhance feature extraction networks' generalization capability via cross-task self-supervised learning, and distinctly separate features under different operating conditions of the same object. By incorporating uncertainty consistency regularization and a sample filtering mechanism, the framework strengthens out-of-distribution detection. Additionally, Bayesian meta-learning is employed to infer priors, enabling dynamic optimization of classifiers. This framework offers a practical and efficient solution for improving the performance of industrial fault diagnosis models from the perspective of data characteristics.

# 2. Theoretical and Mathematical Modeling of Diagnostic Uncertainty In Data Imbalance Conditions

Diagnostic uncertainty caused by data imbalance arises from various sources, including sample data uncertainty (Liao et al., 2023), diagnostic model uncertainty (Che et al., 2025; Mehimeh et al., 2023; Xu et al., 2024), and out-of-domain distribution uncertainty (Ding et al., 2023; Lin et al., 2022). This section begins by clearly defining the core concepts and challenges associated with these uncertainties, with a focus on how data imbalance exacerbates diagnostic difficulties. The objective of this section is to systematically analyze the impact of data imbalance on diagnostic uncertainty throughout the entire diagnostic process, bridge theoretical gaps, and provide mathematical quantification methods. This theoretical foundation underpins the improved diagnostic framework proposed in subsequent sections, ensuring a robust approach to addressing the challenges posed by imbalanced data conditions. As illustrated in Figure 1, the theoretical framework integrates data uncertainty quantification, model uncertainty evaluation, calibration strategies, and Bayesian classifier enhancements, providing a systematic and adaptive approach to tackling diagnostic uncertainty and enabling reliable fault diagnosis.

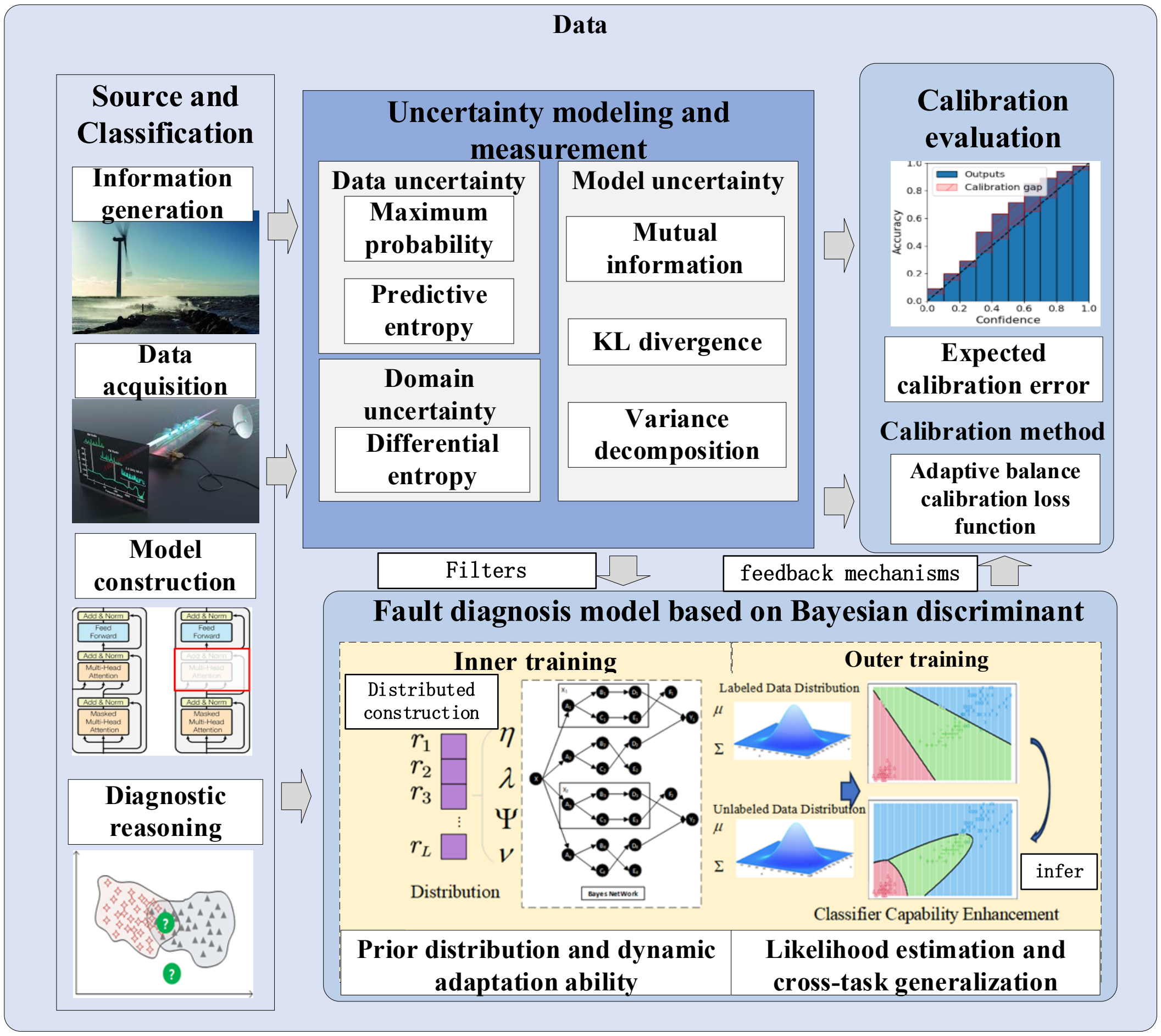

**Fig. 1.** Uncertainty in fault diagnosis of data imbalance and theoretical framework of fault diagnosis

## 2.1. *Comprehensive Uncertainty Modeling and Quantification Framework*

With the widespread application of deep learning in fault diagnosis, data-driven models face critical challenges under data imbalance and out-of-distribution (OOD) conditions. 1) Data imbalance leads to diagnostic biases, as majority-class samples dominate model optimization, causing overfitting to or neglect of minority-class samples. Additionally, traditional models struggle to differentiate between high-noise data and unknown anomalies, often resulting in overconfident fault predictions. 2) The inability to recognize OOD samples poses significant risks, as diagnostic models frequently encounter previously unseen fault modes or environmental conditions in real-world applications. 3) Such models often suffer from poor generalization, limiting their adaptability to cross-task or cross-environment scenarios, thereby restricting their practical utility. Addressing these issues is critical for industrial applications, yet most data-driven fault diagnosis models rely on softmax as the final layer to treat fault diagnosis as a classification problem. Consequently, the uncertainty caused by data imbalance is often quantified directly using softmax outputs, which have notable limitations: 1) they fail to distinguish between cognitive and domain uncertainty, hindering effective evaluation of diagnostic impacts; 2) they are highly sensitive to minor logit fluctuations, leading to distorted quantification; and 3) they struggle to produce uniform outputs for OOD samples, often exhibiting overconfidence under imbalanced conditions. Therefore, to enhance the expressive capacity of diagnostic models in imbalanced and complex scenarios, it is essential to construct a more comprehensive uncertainty modeling and quantification framework across data(2.1.1 Sample Uncertainty), model(2.1.2 Diagnostic Model Uncertainty), and domain levels(2.1.3 Out-of-Domain Generalization Uncertainty).

### *2.1.1. Sample Uncertainty*

**Source:** Sample data uncertainty arises from two key stages: information generation and data acquisition, challenging the authenticity and representativeness of diagnostic samplesIn the information generation stage, electromechanical products, as complex systems with multiple physical fields and components, inherently exhibit uncertainty in their operational states due to design parameter margins, manufacturing tolerances, and assembly errors during maintenance. These are further amplified by dynamic environmental changes (e.g., temperature, humidity, electromagnetic fields). In the data acquisition stage, uncertainty stems from sensor noise, improper measurement placement causing data distortion or loss, and subjective or biased label generation based on expert knowledge.

**Mathematical Model:** To address sample data uncertainty, a multi-task Bayesian framework with variational inference is adopted, decomposing uncertainty into aleatoric uncertainty (inherent data randomness) and epistemic uncertainty (model limitations and data insufficiency). The overall predictive distribution is defined as (1):

$$p(\mathbf{y}^*|\mathbf{x}^*) = \int p(\mathbf{y}^*|\mathbf{x}^*,\theta)p(\theta|\mathcal{D})d\theta \tag{1}$$

Here, $p(\mathbf{y}^*|\mathbf{x}^*,\theta)$ captures the aleatoric uncertainty, reflecting the influence of inherent randomness on model predictions; $p(\theta|\mathcal{D})$ represents the epistemic uncertainty, modeling the impact of limited training data on the model parameters $\theta$.

In multi-task diagnostic scenarios, it is necessary to construct shared and task-specific parameters across tasks. For a given task set $\mathcal{J} = \{T_1,T_2,...,T_m\}$, each task $T_i$ is associated with its dataset $\mathcal{D}_k = \{(\mathbf{x}_k,\mathbf{y}_k)\}_{k=1}^{n}$. The overall parameterization of task-specific and shared parameters is expressed as (2):

$$p(\{\theta_k\}_{k=1}^{m}|\mathcal{D},\phi) \propto \prod_{k=1}^{m} p(\mathcal{D}_k|\theta_k)p(\theta_k|\phi) \tag{2}$$

Where, $\phi$ shared parameter across tasks, representing the global prior distribution; $p(\theta_k|\phi)$ reflects the posterior distribution of task-specific parameters, conditioned on the shared parameters $\phi$. To optimize the parameters, the Evidence Lower Bound (ELBO) is used, defined as (3):

$$\mathcal{L}(\phi,\{\theta_k\}) = \sum_{k=1}^{m} \mathbb{E}_{q(\theta_k)}[\log p(\mathcal{D}_k|\theta_k)] - \mathrm{KL}(q(\theta_k) \parallel p(\theta_k|\phi)) \tag{3}$$

Where, KL $(\cdot)$ : The Kullback-Leibler divergence measures the discrepancy between the approximate posterior $q(\theta_k)$ and the true posterior $p(\theta_k|\phi)$.

This approach facilitates multi-task learning by effectively balancing shared and task-specific parameters, while enhancing uncertainty quantification, thereby improving the adaptability of fault diagnosis models to diverse operational conditions.

**Quantification:**Sample data uncertainty is further decomposed into aleatoric and epistemic components, quantified as follows:

**Aleatoric Uncertainty:** Aleatoric uncertainty is quantified using the entropy of predictive distributions:

$$\text{Aleatoric @ rtainty } = \mathbb{E}_{p(y|\mathbf{x})}[-\log p(y|\mathbf{x})], \tag{4}$$

For classification tasks, this can be simplified to:

$$H(p)\downarrow -\sum_{k} p_k \log p_k, \tag{5}$$

Where, $p_k$ The predicted probability for class $k$. $H(p)$ : Measures the spread of the predicted probabilities, with higher entropy indicating greater uncertainty.

**Epistemic Uncertainty:** Epistemic uncertainty reflects the impact of limited data and model assumptions, estimated through Bayesian posterior modeling of parameters. By quantifying these uncertainties, the framework enhances diagnostic robustness and supports downstream optimization tasks.

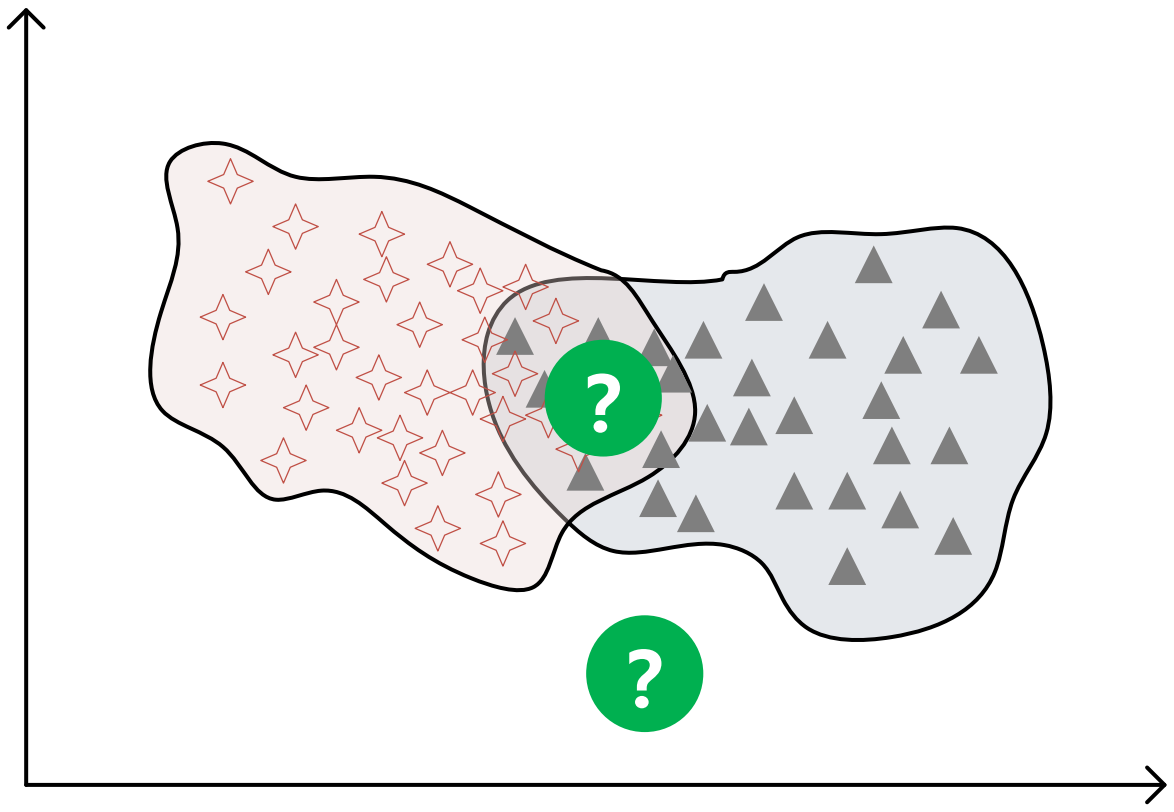

**Fig. 2.** Two different conditions that cause high prediction entropy

### *2.1.2. Diagnostic Model Uncertainty*

**Source:** Model uncertainty (Epistemic Uncertainty) arises from the design and implementation stages of the modeling process, including: **Model Architecture Design**, Structural choices, such as network layers, activation functions, and regularization, directly impact adaptability and generalization, with poor designs underperforming on minority classes. **Hyperparameter Settings,** Factors like learning rate, optimization algorithm, and initialization influence parameter optimization paths, affecting performance in complex tasks. **Evaluation and Selection,** Incomplete evaluations, particularly under limited data, can lead to parameter instability, with stopping criteria and validation set strategies playing critical roles.

**Mathematical Model:** The integration of Bayesian inference with KL divergence provides systematic theoretical support for uncertainty modeling and quantification under data imbalance. By combining Bayesian inference with KL divergence, the framework offers theoretical tools to model uncertainty while adapting predictions to reflect parameter uncertainty. For diagnostic tasks, the predictive uncertainty is expressed as (6):

$$p(\mathbf{y}^*|\mathbf{x}^*,\mathcal{D}) = \int p(\mathbf{y}^*|\mathbf{x}^*,\theta)p(\theta|\mathcal{D})d\theta \tag{6}$$

Where, $\mathcal{D}$ : Represents the training dataset; $p(\theta|\mathcal{D})$ : Denotes the posterior distribution of the parameters $\theta$;$p(\mathbf{y}^*|\mathbf{x}^*,\theta)$ : Reflects the predictive uncertainty given model parameters $\theta$.

KL divergence is employed to measure the difference between the predicted distribution $p(\mathbf{y}^*|\mathbf{x}^*,\mathcal{D})$ and the observed distribution $p(\mathbf{y}^*|\mathbf{x}^*,\theta)$, defined as (7):

$$KL(p(\mathbf{y}^*|\mathbf{x}^*,\mathcal{D}) \parallel p(\mathbf{y}^*|\mathbf{x}^*,\theta)) = \int p(\mathbf{y}^*|\mathbf{x}^*,\mathcal{D})\log\frac{p(\mathbf{y}^*|\mathbf{x}^*,\mathcal{D})}{p(\mathbf{y}^*|\mathbf{x}^*,\theta)}d\mathbf{y}^* \tag{7}$$

To further refine the quantification of uncertainty, the expected KL divergence over the posterior distribution of $\theta$ is defined as (8):

$$\mathbb{E}_{p(\theta|\mathcal{D})}\left[KL\left(p\left(\mathbf{y}^*|\mathbf{x}^*,\mathcal{D}\right) \parallel p\left(\mathbf{y}^*|\mathbf{x}^*,\theta\right)\right)\right] \tag{8}$$

This expectation quantifies the inconsistency between the model predictions and posterior parameter distributions, providing a direct measure of model uncertainty. By integrating Bayesian inference and KL divergence, the uncertainty is further expressed as (9):

$$H\left(p\left(\mathbf{y}^*|\mathbf{x}^*,\mathcal{D}\right)\right) = H\left(p\left(\mathbf{y}^*|\mathbf{x}^*,\theta\right)\right) + MI\left(\mathbf{y}^*,\mathbf{x}^*\right) \tag{9}$$

Where, $H\left(p\left(\mathbf{y}^*|\mathbf{x}^*,\theta\right)\right)$: Measures predictive uncertainty arising from parameter variability. $MI\left(\mathbf{y}^*,\mathbf{x}^*\right)$ : Represents mutual information, reflecting the dependence between predictions and data. It evaluates the degree to which parameter uncertainty impacts predictions, offering a more comprehensive representation of uncertainty.

**Quantification:** Mutual information **(MI)** captures the uncertainty reduction when additional information is provided. It is defined as (10):

$$MI\left(\mathbf{y}^*,\mathbf{x}^*\right) = H\left(p\left(\mathbf{y}^*|\mathbf{x}^*,\theta\right)\right) - \mathbb{E}_{p(\theta|\mathcal{D})}\left[H\left(p\left(\mathbf{y}^*|\mathbf{x}^*,\theta\right)\right)\right] \tag{10}$$

Here, $H\left(p\left(\mathbf{y}^*|\mathbf{x}^*,\theta\right)\right)$ Measures predictive uncertainty from parameter variability; $\mathbb{E}_{p(\theta|\mathcal{D})}\left[H\left(p\left(\mathbf{y}^*|\mathbf{x}^*,\theta\right)\right)\right]$ Captures posterior-averaged uncertainty, reflecting the uncertainty over all parameters. By combining the above equations, this formulation encapsulates the predictive uncertainty, balancing the observed variability and posterior uncertainty.

### *2.1.3. Out-of-Domain Generalization Uncertainty*

**Source:** Out-of-Domain (OOD) uncertainty arises when inference involves data samples that deviate from the training or operational data distribution. This is often caused by incomplete or biased training data coverage. For instance, long-term operations or new environmental conditions may lead to distributional shifts. Real-world tasks, due to their openness and variability, make handling OOD uncertainty critical. If not addressed, OOD uncertainty can lead to significant errors in decision-making, especially in safety-critical scenarios like autonomous driving or fault diagnosis.

**Mathematical Model:** OOD uncertainty is commonly modeled using Dirichlet Prior Network (DPN), which captures predictive uncertainty through the posterior distribution. For a multi-class task, the distribution is defined as (11):

$$p(\mu|\alpha) = \frac{\Gamma\left(\sum_{k=1}^{K}\alpha_k\right)}{\prod_{k=1}^{K}\Gamma(\alpha_k)}\prod_{k=1}^{K}\mu_k^{\alpha_k-1}, \tag{11}$$

where $\mu_k$ represents the predictive probability for class $k_r\alpha = [\alpha_1,\alpha_2,\ldots,\alpha_K]$ are the Dirichlet parameters, and $K$ is the number of classes. The parameter $\alpha_k$ reflects the confidence in predictions. DPN uses this to capture OOD uncertainty effectively.

**Quantification:** OOD uncertainty is typically measured using entropy, which quantifies the uncertainty in the predictive distribution. For an input sample $(x^*,\mu)$, entropy is calculated as (12):

$$H(p(\mu|\alpha)) = -\sum_{k=1}^{K}\mu_k\log(\mu_k), \tag{12}$$

where $H(p(\mu|\alpha))$ measures the uncertainty level. Higher entropy indicates greater uncertainty, especially for OOD samples, where the predictive distribution becomes more uniform. This property allows DPN to detect OOD data and improve model robustness.

## *2.2. Uncertainty-Based Model Calibration*

Uncertainty calibration aims to adjust the confidence of model outputs to align with diagnostic accuracy, addressing the issue of overconfidence in deep neural networks, particularly under data imbalance conditions. The Adaptive Balanced Calibration Loss function combines regularization and post-processing optimization strategies, dynamically adjusting the weights for different classes and sample difficulties during training. On one hand, regularization guides parameter updates by assigning higher weights to minority classes and hard-to-classify samples, mitigating the impact of data imbalance. On the other hand, post-processing optimizes confidence outputs to bring them closer to actual predictive probabilities. This approach balances calibration and classification performance, significantly improving diagnostic accuracy and uncertainty calibration while enhancing model reliability. Moreover, its implementation is simple, requiring no complex modifications, as performance improvements are achieved solely by optimizing the loss function. The model calibration process can be summarized into the following key steps:

**Metric Evaluation:** The model's reliability is assessed through two metrics: confidence ( conf) and accuracy (acc). Confidence represents the average predicted probability of samples, while accuracy denotes the actual classification correctness. The consistency between $\mathrm{conf}(\mathrm{b_m})$ and $\mathrm{acc}(\mathrm{b_m})$, as visualized through a Reliability Diagram, determines the calibration effectiveness.

**Calibration Issues in Classification Tasks:** To address overconfidence or underconfidence in the model, the calibration process identifies different states of reliability, such as "underconfident," "overconfident," and "well-calibrated." This helps pinpoint optimization directions. Given an input sample $x$, the model outputs the probability of its health state as $p=[p_1,p_2,...,p_K]$. For calibration, the model's output confidence $p_c$ should satisfy $P(y=\arg\max p)=p_c$, where $y$ represents the actual label. To analyze calibration, the samples are divided into $M$ intervals based on confidence $p_i$, denoted as $b_1,b_2,...,b_M$. For each interval $b_m$, the average confidence $\mathrm{conf}(b_m)$ and accuracy $\mathrm{acc}(b_m)$ are computed as (13):

$$\begin{aligned}\mathrm{conf}(b_m)&=\frac{1}{|b_m|}\sum_{i\in b_m}p_i,\\ \mathrm{acc}(b_m)&=\frac{1}{|b_m|}\sum_{i\in b_m}I(\dot{y}_i=y_i),\end{aligned}\tag{13}$$

where $\dot{y}_i$ and $p_i$ denote the predicted label and confidence for the $i$-th sample, respectively, $y_i$ is the true label, and $I$ is an indicator function that equals 1 if the prediction is correct and 0 otherwise. $|b_m|$ represents the number of samples within the interval $b_m$.

The Reliability Diagram visually compares $\mathrm{conf}(b_m)$ and $\mathrm{acc}(b_m)$ for each interval. Perfect calibration is achieved when $\mathrm{conf}(b_m)=\mathrm{acc}(b_m)$ for all intervals, indicating the model's predictions are well-calibrated. If $\mathrm{conf}(b_m)\neq\mathrm{acc}(b_m)$, the gap reflects the calibration error. In a wellcalibrated model, $\mathrm{conf}(b_m)$ aligns with $\mathrm{acc}(b_m)$, and the reliability diagram displays a diagonal line.

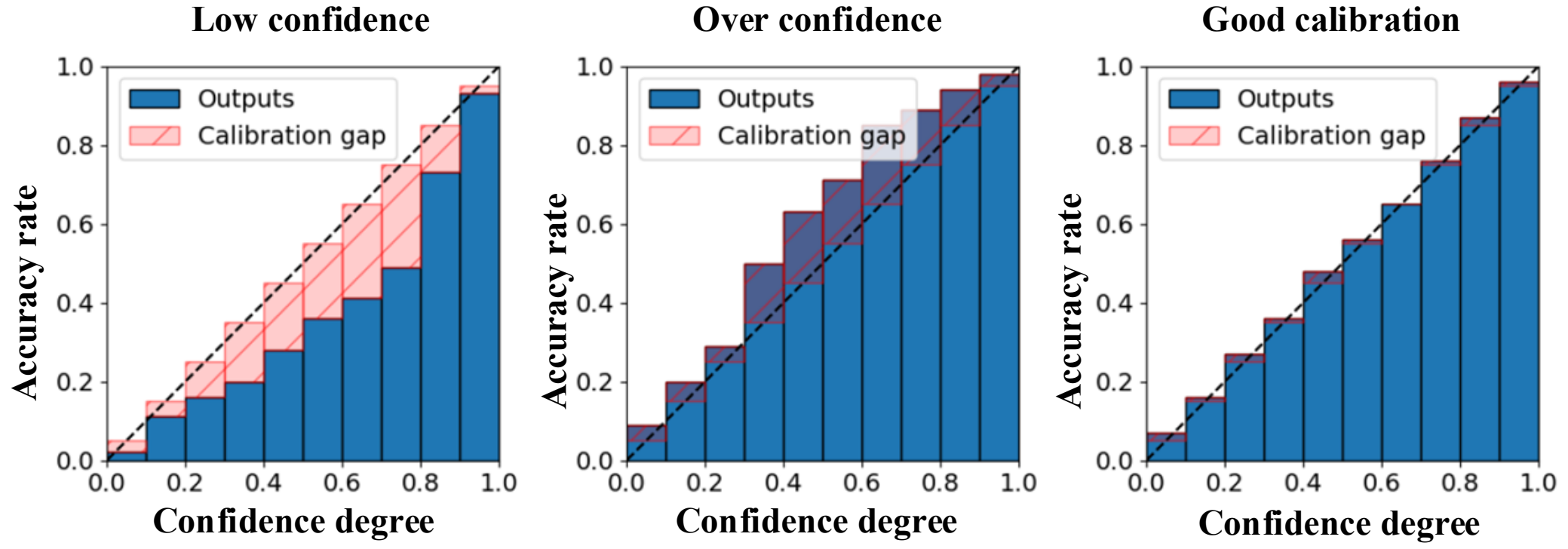

**Fig. 3.** Reliability diagram for different calibration conditions

**Adaptive Balanced Calibration Strategy:** The Adaptive Balanced Calibration Loss (aBCE) is designed to address both class imbalance and uncertainty calibration. It constrains prediction bias $(\mathrm{conf}(N_c))$ and corrects the gap between confidence and actual accuracy $(\mathrm{acc}(b_m)-\mathrm{conf}(b_m))$ to ensure calibration effectiveness.

The adaptive loss function is defined as (14):

$$aBCE=\frac{|b|}{K^2}\sum_{c=1}^{K}(1-\mathrm{conf}(N_c))^v+\frac{1}{M}\sum_{m=1}^{M}(\mathrm{acc}(b_m)-\mathrm{conf}(b_m)),\tag{14}$$

where $N_c$ represents the prediction bias for a class, and $\text{acc}(b_m) - \text{conf}(b_m)$ measures the calibration error.

Joint Optimization: The final objective combines the classification loss $(L(\theta))$ and calibration loss $(aBCE$ $)$ as (15):

$$L_{BCE} = \frac{1}{|B|} \sum_{i \in B} (L(\theta) + \beta \cdot aBCE) \tag{15}$$

where $L(\theta)$ is the base classification loss, and $aBCE$ is the calibration term.

This process dynamically adjusts loss weights to optimize confidence and classification performance, significantly enhancing the model's reliability and diagnostic accuracy under imbalanced data conditions.

### *2.3. Fault Diagnosis Framework Based on Bayesian Classifiers and Uncertainty*

The Bayesian diagnostic framework focuses on quantifying uncertainty by integrating prior knowledge and observed data. It processes data streams, selects samples, and trains models to improve diagnostic accuracy and reliability. Uncertainty Quantification: The diagnostic objective is defined as (16):

$$J(f(y|x)) = \sum_c P(y = c|x) \log P(y = c|x) \tag{16}$$

where $P(y = c|x)$ represents the predicted probability for class $c$. Entropy is used to evaluate and adjust prediction uncertainty.

Posterior Distribution Adjustment: The posterior probability is computed as (17):

$$P(G|D) = \frac{P(D|G)P(G)}{P(D)} \tag{17}$$

where $P(D|G)$ is the likelihood, and $P(G)$ is the prior. This combines prior knowledge and data-driven likelihood estimates to refine predictions.

**Calibration Evaluation:** The Expected Calibration Error (ECE) quantifies the alignment of confidence and accuracy (18):

$$\text{ECE} = \sum_{m=1}^{M} \frac{|b_m|}{n} |\text{acc}(b_m) - \text{conf}(b_m)|, \tag{18}$$

where $\text{acc}(b_m) = \frac{1}{|b_m|} \sum_{i \in b_m} I(\hat{y}_i = y_i)$ is the accuracy, $\text{conf}(b_m) = \frac{1}{|b_m|} \sum_{i \in b_m} p_i$ is the confidence, $|b_m|$ is the sample count in bin $b_m$, and $n$ is the total number of samples.

This framework dynamically adjusts model predictions to handle uncertainty and overconfidence, especially for out-of-distribution (OOD) samples. By integrating Bayesian principles with calibration metrics, it enhances the robustness and reliability of fault diagnosis.

## 3. Methods

The model (UBMF in short) implementation is divided into three stages. In the first stage, it leverages unlabeled historical data by designing a metric network that can autonomously adjust for domain differences across tasks. Using a transductive meta-learning approach, the model efficiently propagates pseudo-labels for unlabeled samples, balancing the class distribution within each small batch to mitigate class imbalance during learning. In the second stage, through reasonable pretext tasks and optimization objectives, a meta self-supervised learning method is employed to guide the feature extraction network in capturing the internal structure of the data, enhancing generalization ability. Domain-specific data perturbations, categorized as weak and strong based on their impact on diagnostic reasoning, are injected into the self-supervised feature extraction process. Knowledge is incorporated through consistency regularization of the predictions, improving feature extraction. In the third stage, Bayesian meta-learning is applied, where prior knowledge from historical tasks is used to infer a multivariate normal distribution in the secondary classifier. This enables the model to uncover diagnostic knowledge and effectively fit the classifier, resulting in trustworthy diagnostic outputs. The overall methodology is illustrated in Figure 4.

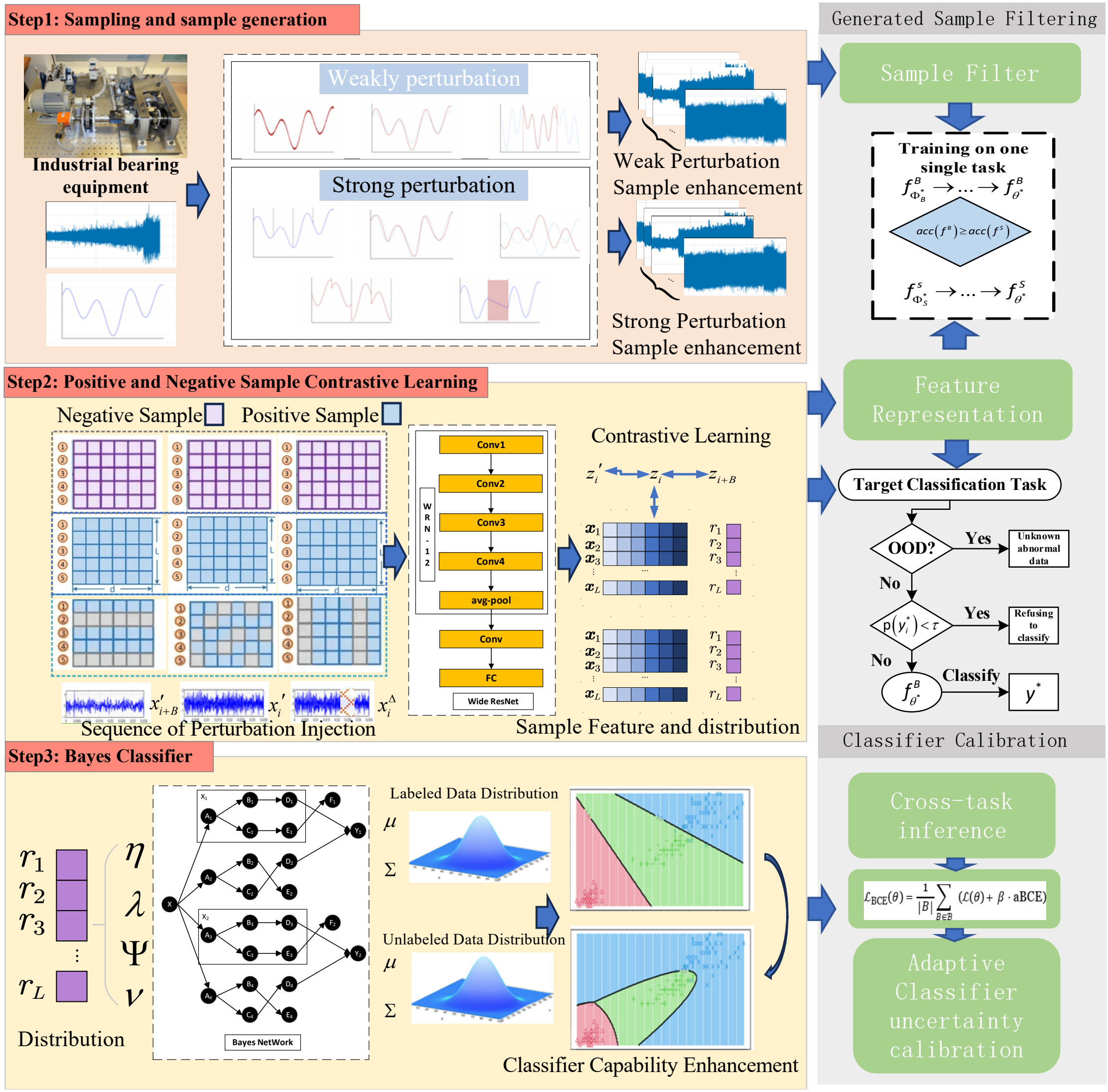

**Fig. 4.** Enhancement of Fault Diagnosis Model Capabilities Based on Extraction and Embedding

## *3.1. Data Perturbation Injection Based on Prior Knowledge*

Consistency regularization, leverages Bayesian prior meta-knowledge to design and evaluate domain-guided weak(3.1.1) and strong(3.1.2) perturbations, refining diagnostic predictions, enhancing uncertainty quantification, and iteratively improving reasoning robustness and prior knowledge.

### *3.1.1. Weak Perturbation*

Weak perturbation methods, such as noise jitter, homogeneous scaling, and splicing resampling (Figure 5), minimally impact diagnostic reasoning and preserve model confidence in electromechanical monitoring data.

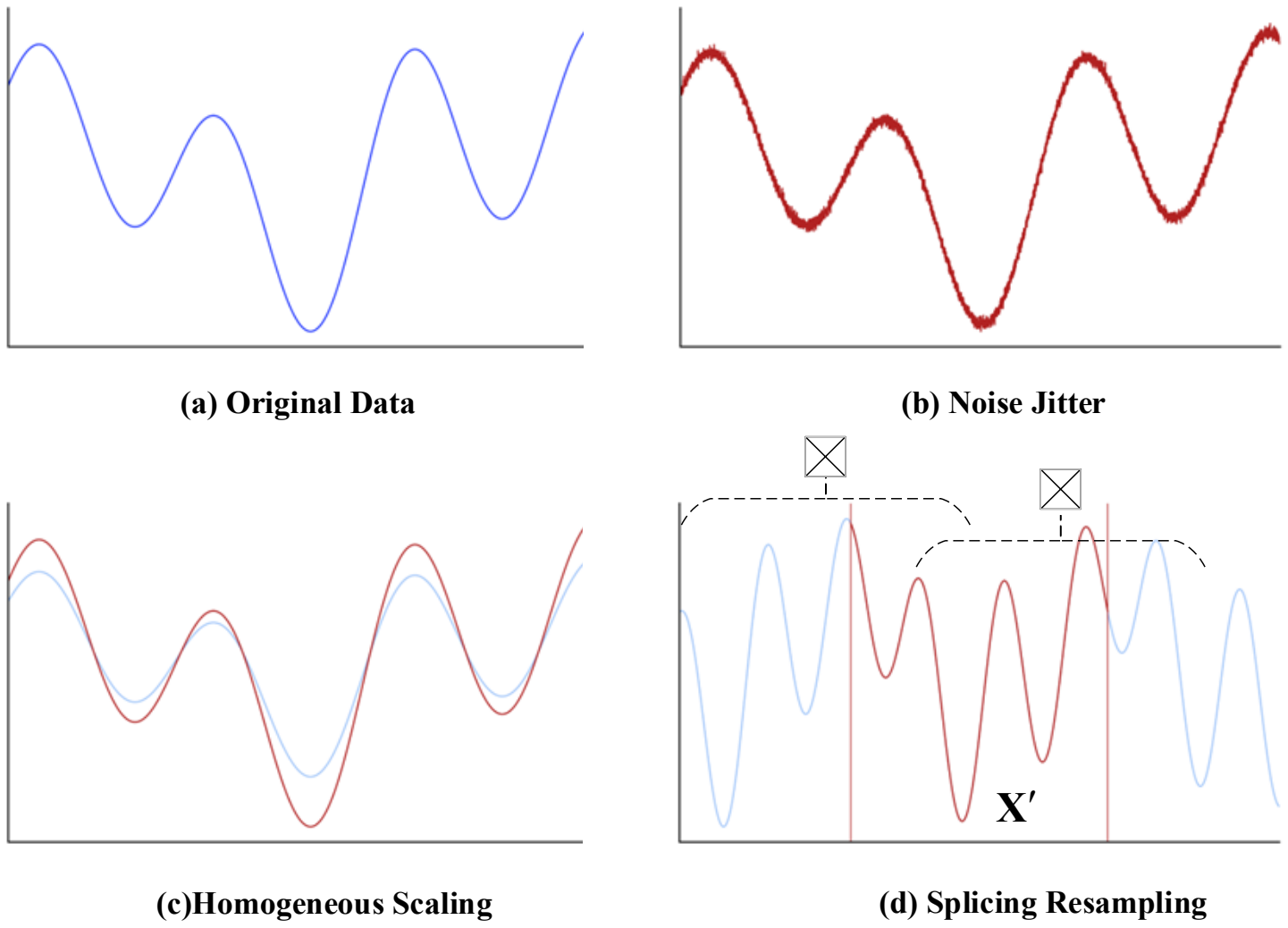

**Fig. 5.** Weak Perturbation Diagram

**Noise jitter** introduces small random noise $\epsilon$ to the original data, formulated as:

$$\mathbf{X}' = [x_1 + \epsilon_1, x_2 + \epsilon_2, \ldots, x_T + \epsilon_T] \tag{19}$$

This method does not alter the fault mode of the data, ensuring the stability of diagnostic reasoning, and is thus categorized as a weak perturbation.

**Homogeneous scaling** applies the same scaling factor $\alpha > 0$ to each dimension of the data, with $\alpha$ sampled from a normal distribution. The formula is:

$$\mathbf{X}' = [\alpha x_1, \alpha x_2, \ldots, \alpha x_T] \tag{20}$$

This method adjusts the data distribution slightly without significantly affecting model confidence, making it suitable for various sample instances.

**Splicing resampling** extracts segments from high-confidence pseudo-labeled samples and labeled samples, combining them using a sliding window to measure similarity. If the similarity $D(X^m, X^n)$ exceeds a threshold $\tau$, the segments are spliced, as formulated:

$$S_{\text{new}} = [x_1^m, \ldots, x_i^m, x_{i+1}^n, \ldots, x_T^n], \text{ where } D(X^m, X^n) \geq \tau \tag{21}$$

A new sample $\mathbf{X}'$ is then generated from $S_{\text{new}}$, ensuring consistent diagnostic confidence in the resulting data.

### *3.1.2. Strong Perturbation*

Strong perturbation methods, such as amplitude distortion, time alignment, rotation, slice rearrangement, interpolation generation, and generative perturbation injection (Figure 6), significantly impact diagnostic reasoning and confidence while preserving sample class.

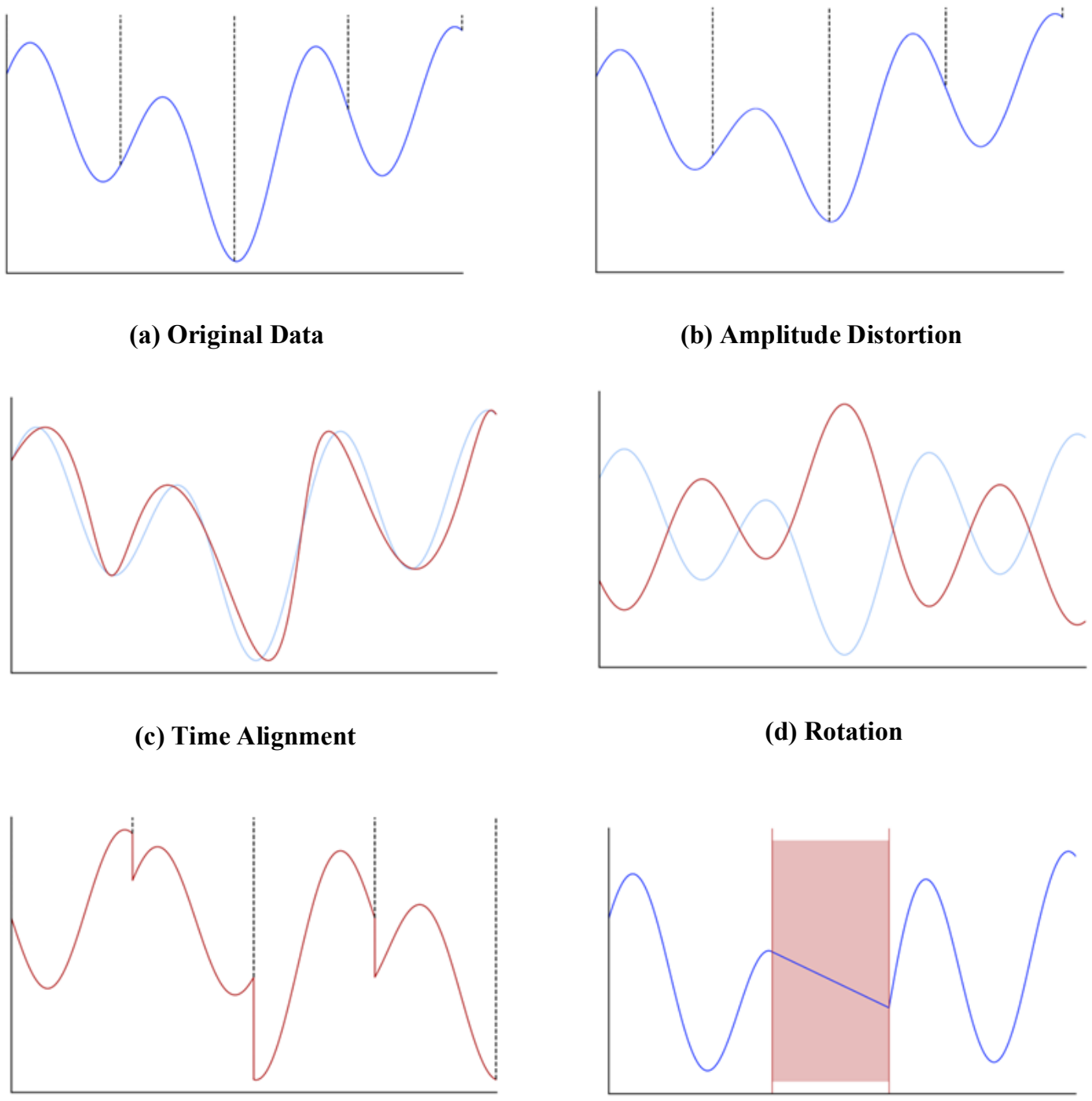

**Fig. 6.** Basic Strong Perturbation Diagram

**Amplitude Distortion:** Applies variable scaling to different dimensions of the sample using cubic spline interpolation to define phase segmentation (22):

$$\mathbf{X}'(\mathbf{a}) = [\alpha_1 x_1, \alpha_2 x_2, \ldots, \alpha_T x_T],\ \mathbf{a} = S(\mathbf{v}) \tag{22}$$

**Time Alignment:** Modifies the monitoring data by stretching or compressing along the time dimension using alignment nodes, similar to amplitude distortion (23):

$$\mathrm{X}'(\tau) = [x_{\tau(1)}, x_{\tau(2)}, \ldots, x_{\tau(T)}] \tag{23}$$

**Rotation:** Transforms the monitoring sequence using a rotation matrix $\mathbf{R}$, often used for spatial feature parametrization (24):

$$\mathbf{X}'(\mathbf{R}) = [R_1 x_1, R_2 x_2, \ldots, R_T x_T] \tag{24}$$

**Slice Rearrangement:** Divides the monitoring data into fixed or variable-length segments and rearranges them to form a new sample (25):

$$\mathbf{X}(\mathbf{w}) = [x_i, \ldots, x_j, x_k, \ldots, x_w] \tag{25}$$

Here, $i,j,k$ are segment starting points, and $w$ is the segment length.

**Random Masking:** Randomly replaces positions in the sample with Gaussian noise, linear interpolation, or zeros to simulate controlled errors (26):

$$\mathbf{X}' = [x_1, x_2, \ldots, x_i, \epsilon_1, \epsilon_2, \ldots, x_T] \tag{26}$$

### *3.1.3. Perturbation Injection Method*

Perturbation injection methods introduce controlled variability into data through feature reconstruction, interpolation, and generative models, enabling robust diagnostic reasoning and enhanced adaptability while maintaining class consistency, as shown in Fig. 7.

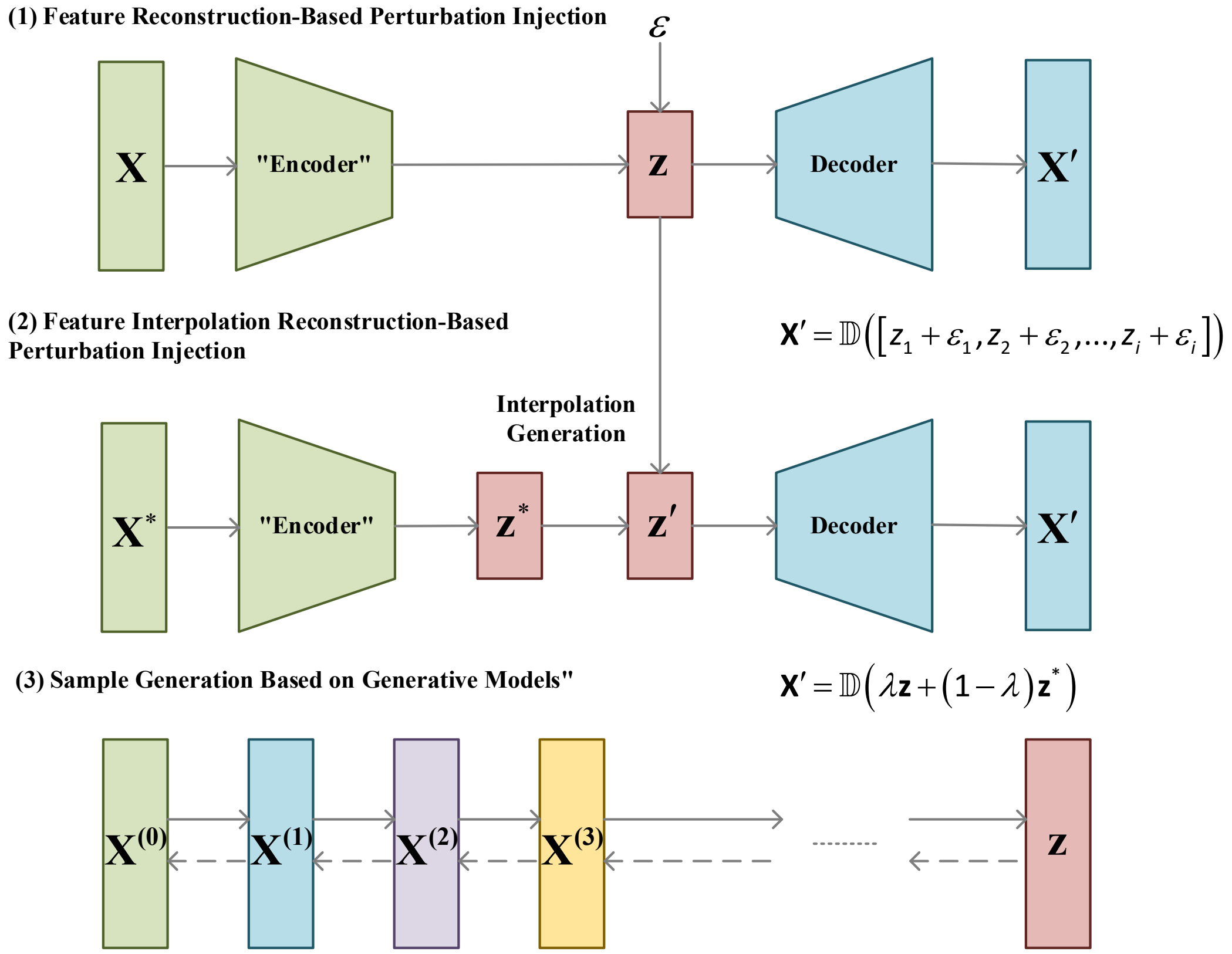

**Fig. 7.** Generative Strong Perturbation Injection Diagram

**Feature Reconstruction-Based Perturbation Injection:**

Using autoencoders or meta-learning networks, weak perturbations (e.g., Gaussian noise) are added to the latent feature code, which is reconstructed into the original sample form:

$$X' = \mathcal{D}(z'(\epsilon)) = \mathcal{D}([z_1 + \epsilon_1, z_2 + \epsilon_2, ..., z_i + \epsilon_i]). \tag{27}$$

Here, $\mathcal{E}$ and $\mathcal{D}$ are the encoder and decoder networks, and $z = \mathcal{E}(X)$ represents the feature latent code. Strong perturbations emerge from amplified weak perturbations, requiring reliable reconstruction methods.

**Feature Interpolation-Based Perturbation Injection:**

Generate new samples by interpolating between the latent codes of high-confidence unlabeled samples and labeled samples of the same class:

$$X' = \mathcal{D}(z'(\lambda)) = \mathcal{D}(\lambda z + (1-\lambda) z^*) \tag{28}$$

where $z$ and $z^*$ are the latent codes, and $\lambda$ controls the interpolation. This method ensures domain alignment and requires expert knowledge for optimal interpolation.

**Sample Generation-Based Perturbation Injection:**

Utilize generative models (e.g., VAEs, GANs, Diffusion Models) to produce high-quality perturbed samples. These models replace original data with generated samples of the same class, enabling flexible and controlled strong perturbation injection.

## *3.2. Cross-task Self-supervised Feature Extraction Enhancement*

Section 3.1 introduces perturbation-based sample generation to enrich data diversity, while Section 3.2 leverages cross-task contrastive learning to capture robust, transferable features. By aligning domain-specific knowledge with global feature spaces, this approach not only mitigates data imbalance effects but also enhances feature representation and generalization to unseen tasks.

### *3.2.1. Learnable Metric Network*

To mitigate the interference of an imbalanced, unlabeled sample space during selfsupervised feature extraction, this section proposes a meta-learnable encoder network to adjust the Euclidean distance metric dynamically. The encoder $E_{\phi}(X_i)$ is specifically designed to modify the Euclidean distance in (29):

$$d(X_i,X_j) = \sqrt{\sum_{k=1}^{n}\left(X_i^k - X_j^k\right)^2} \tag{29}$$

This standard metric calculates the straight-line distance between two vectors $X_i$ and $X_j$ in an $n$-dimensional feature space.The modified distance metric $d_{\phi}$ is defined as (30):

$$d_{\phi}(X_i,X_j) = \left\| \frac{X_i}{\| X_i \| \cdot E_{\phi}(X_i)} - \frac{X_j}{\| X_j \| \cdot E_{\phi}(X_j)} \right\| \tag{30}$$

The contribution weight $\tilde{\delta}_{j,c}$ of the unlabeled sample is updated to (31):

$$\tilde{\delta}_{j,c} = \frac{\exp(-d_{\phi}(f_{\theta}(\tilde{X}_j) - p_c))}{\sum_{c'}\exp(-d_{\phi}(f_{\theta}(\tilde{X}_j) - p_{c'}))} \tag{31}$$

The probability vector of the test sample $X^*$ is updated to (32):

$$P(c|X^*,\{p_c\}) = \frac{\exp(-d_{\phi}(f_{\theta}(X^*) - p_c))}{\sum_{c'}\exp(-d_{\phi}(f_{\theta}(X^*) - p_{c'}))}. \tag{33}$$

As the scale adjustment factor, the encoder $E_{\phi}(X_i)$ is trained every T meta-learning epochs, and its loss function is designed as (34):

$$\mathrm{L}_d(\theta,\phi) = \frac{1}{|\mathrm{D}|}\sum_{(\tilde{X},\tilde{y}) \in}\{d_{\phi}(f_{\theta}(\tilde{X}),p_c^{(T)}) + \sum_{c'=1}^{C}\exp(-d_{\phi}(f_{\theta}(\tilde{X}),p_{c'}^{(T)}))\}, \tag{34}$$

Here, $p_c^{(T)}$ represents the class prototype of class $c$ in the T-th epoch. The learnable scale factor encoder $E_{\phi}(X_i)$ adjusts inter-domain differences, enabling more reasonable class prototype updates from unlabeled samples across domains. A shallow encoder achieves this efficiently without compromising transductive learning performance.

### *3.2.2. Small Batch Self-Supervised Loss Design*

To preserve cross-domain information while addressing uncertainty, this method maximizes mutual information between the original sample $\mathbf{X}$ and its feature vector $\mathbf{z}_l$ promoting a more dispersed feature distribution across domains. Direct computation of mutual information is infeasible, so a contrastive learning approach is adopted. Augmented or perturbed versions of the same sample, $\mathbf{X}$ and $\mathbf{X}'$, are treated as positive pairs, while different samples are treated as negative pairs. The objective is to align positive pairs closely and uniformly separate negative pairs.

Weak perturbations $\epsilon$ are applied twice to each unlabeled sample $\mathbf{X}$, generating positive pairs $(\mathbf{X},\mathbf{X}')$. The feature extraction network $f_{\theta}$ processes these samples, obtaining feature vectors $\mathbf{z}_i$ and $\mathbf{z}_i'$. The self-supervised loss is defined as (35):

$$\mathcal{L}_{\mathrm{SSL}} = -\frac{1}{B}\sum_{i=1}^{B}\frac{\mathbf{z}_i \cdot \mathbf{z}_i'}{\tau \| \mathbf{z}_i \| \ \| \mathbf{z}_i' \|} + \frac{1}{2B}\sum_{i=1}^{B}\log\left(\sum_{j=1}^{2B} e^{\mathbf{z}_i \cdot \mathbf{z}_j/\tau}\right) \tag{35}$$

Here, $B$ is batch size;$\tau$, Temperature hyperparameter controlling distribution sharpness; $\mathbf{z}_j$ ,Feature vectors of negative samples. In small-batch scenarios, episodes may suffer from insufficient information. To address this, the loss for small-batch size $B$ is modified as (36):

$$\mathcal{L}_{\mathrm{SSL}} = -\frac{1}{B}\sum_{i=1}^{B}\frac{\mathbf{z}_i \cdot \mathbf{z}_i'}{\tau \parallel \mathbf{z}_i \parallel \parallel \mathbf{z}_i' \parallel} + \log\left(\sum_{j=1}^{2B} e^{\mathbf{z}_i \cdot \mathbf{z}_j/\tau}\right) \tag{36}$$

This modification balances batch information, ensuring effective alignment and uniformity during self-supervised training.

### 3.2.3. Knowledge Embedding through Perturbation Injection

To enhance diagnostic effectiveness, this section introduces a consistency regularization method that embeds prior knowledge into the training process. By applying regularization on the uncertainty of model predictions under varying levels of data perturbations, it improves the reliability of extracted features for cross-task diagnostics.

After $T$ epochs of cross-task self-supervised training, the feature extraction network $f_\theta$ computes the class prototype $p_c$ for each class $c$ using labeled samples $\mathbf{X}_i$. For an unlabeled sample $\mathbf{X}$, weak perturbations $\delta$ are applied to obtain pseudo-label vectors $P(y|\delta(\mathbf{X}))$, as (37):

$$P(y|\delta(\mathbf{X})) = \mathrm{softmax}\left(f_\theta(\delta(\mathbf{X}))\right) \tag{37}$$

Unlabeled samples with high-confidence pseudo-labels, denoted as $\grave{\mathbf{X}}_j$, are selected. Strong perturbations $\Delta$ are then applied to these samples, and the pseudo-labels from weak perturbations serve as soft labels. The supervised loss $\mathcal{L}_s$ is computed as (38):

$$\mathcal{L}_s = \frac{1}{|\grave{\mathbf{D}}|}\sum_j \mathbb{I}\left(P_{\max}\left(\delta\left(\grave{\mathbf{X}}_j\right)\right) > \tau\right) \cdot \mathcal{H}\left(P\left(y|\delta\left(\grave{\mathbf{X}}_j\right)\right), P\left(y|\Delta\left(\grave{\mathbf{X}}_j\right)\right)\right) \tag{38}$$

Here, $\grave{\mathbf{D}}$,Selected unlabeled samples with high-confidence pseudo-labels; $P_{\max}$, Maximum probability of pseudo-label under weak perturbations; $\tau$, Confidence threshold hyperparameter; $\mathcal{H}$, Cross-entropy function. The final loss combines the pseudo-supervised loss $\mathcal{L}_s$ with the self-supervised loss $\mathcal{L}_{\mathrm{SSL}}$ :

$$\mathcal{L}_{\mathrm{total}} = \mathcal{L}_{\mathrm{SSL}} + \lambda\mathcal{L}_s \tag{39}$$

This method leverages consistency regularization to guide feature extraction, ensuring robust and transferable feature representations for diagnostic tasks.

## 3.3. Classifier Enhancement Based on Bayesian Meta-Knowledge Extraction

To enhance cross-task diagnostics and feature extraction, a Quadratic Discriminant Analysis (QDA) classifier is employed for improved fitting, while Bayesian meta-learning integrates prior knowledge of the multivariate normal distribution, mitigating overfitting.

### 3.3.1. Bayesian Quadratic Classifier Design

For a sample $X \in \mathrm{R}^d$, with a multivariate normal distribution $N$, a mean value $\mu \in \mathrm{R}^d$, and a covariance matrix $\Sigma \in \mathrm{R}^{d \times d}$, the basic quadratic discriminant analysis for inferring the health state $k$ is given by:

$$y^* = \underset{k}{\mathrm{argmax}}\left(-\frac{1}{2}(X-\mu_k)^{\mathrm{T}}\Sigma_k^{-1}(X-\mu_k) + \ln(\pi_k)\right). \tag{40}$$

By using the N data points belonging to health state $k$ in the training data, the distribution parameters of health state $k$ are inferred using maximum likelihood (41):

$$\mu_k, \Sigma_k = \mathrm{argmax}\prod_{i=1}^{N} N(X_{k,i}|\mu,\Sigma). \tag{41}$$

To perform Bayesian inference, the conjugate prior of the multivariate normal distribution is chosen: the Normal-inverse-Wishart (NIW) distribution as the prior for the mean and covariance matrix, denoted as $(\mu,\Sigma) \sim NIW(\eta,\lambda,\Psi,\nu)$. The posterior distribution of the parameters is then given by:

$$p\left(\boldsymbol{\mu}_k,\boldsymbol{\Sigma}_k \mid \boldsymbol{X},\boldsymbol{\eta},\lambda,\Psi,\nu\right)=\frac{\prod_{i=1}^{N}\mathcal{N}\left(\boldsymbol{X}_{k,i}\mid\boldsymbol{\mu}_k,\boldsymbol{\Sigma}_k\right)\mathcal{NIW}\left(\boldsymbol{\mu}_k,\boldsymbol{\Sigma}_k\mid\boldsymbol{\eta},\lambda,\Psi,\nu\right)}{\iint\prod_{i=1}^{N}\mathcal{N}\left(\boldsymbol{X}_{k,i}\mid\boldsymbol{\mu},\boldsymbol{\Sigma}\right)\mathcal{NIW}\left(\boldsymbol{\mu}',\boldsymbol{\Sigma}'\mid\boldsymbol{\eta},\lambda,\Psi,\nu\right)d\boldsymbol{\mu}'d\boldsymbol{\Sigma}'},\tag{42}$$

As follows:

$$\begin{gathered}\boldsymbol{\eta}_k=\left(\lambda\boldsymbol{\eta}+N\overline{\boldsymbol{X}}\right)/\left(\lambda+N\right),\quad \lambda_k=\lambda+N,\quad \nu_k=\nu+N,\\ \Psi_k=\Psi+S+\frac{\lambda N}{\lambda+N}\left(\overline{\boldsymbol{X}}-\boldsymbol{\eta}\right)\left(\overline{\boldsymbol{X}}-\boldsymbol{\eta}\right)^{\mathrm{T}},\quad S=\sum_{i=1}^{N}\left(\boldsymbol{X}_i-\overline{\boldsymbol{X}}\right)\left(\boldsymbol{X}_i-\overline{\boldsymbol{X}}\right)^{\mathrm{T}}.\end{gathered}\tag{43}$$

During inference, the parameter distribution is marginalized and expanded using Bayes' theorem, as shown (44):

$$p\left(y^*=k\mid\boldsymbol{X}\right)=\frac{\iint\mathcal{N}\left(\boldsymbol{X}\mid\boldsymbol{\mu}_k,\boldsymbol{\Sigma}_k\right)\mathcal{NIW}\left(\boldsymbol{\mu}_k,\boldsymbol{\Sigma}_k\mid\boldsymbol{\eta}_k,\lambda_k,\Psi_k,\nu_k\right)d\boldsymbol{\mu}_k d\boldsymbol{\Sigma}_k}{\sum_{k=1}^{C}\iint\mathcal{N}\left(\boldsymbol{X}\mid\boldsymbol{\mu}_k,\boldsymbol{\Sigma}_k\right)\mathcal{NIW}\left(\boldsymbol{\mu}_k,\boldsymbol{\Sigma}_k\mid\boldsymbol{\eta}_k,\lambda_k,\Psi_k,\nu_k\right)d\boldsymbol{\mu}_k d\boldsymbol{\Sigma}_k},\tag{44}$$

The above double integral can be converted into the form of a multivariate t-distribution (Zhang et al., 2021), as shown (45):

$$p\left(y^*=k\mid\boldsymbol{X},\boldsymbol{\eta},\lambda,\Psi,\nu\right)=\frac{\mathcal{T}\left(\boldsymbol{X}\mid\boldsymbol{\eta},\frac{\lambda_k+1}{\lambda_k\left(\nu_k-d+1\right)}\Psi_k,\nu_k-d+1\right)}{\sum_{i=1}^{C}\mathcal{T}\left(\boldsymbol{X}\mid\boldsymbol{\eta},\frac{\lambda_k+1}{\lambda_k\left(\nu_k-d+1\right)}\Psi_k,\nu_k-d+1\right)}.\tag{45}$$

Finally, to obtain the above posterior, the parameters of the multivariate t-distribution are the meta-knowledge to be learned across tasks in this section, denoted as $\phi=(\eta,\lambda,\Psi,\nu)$. On the labeled dataset, the log-likelihood of the classifier is given by:

$$\mathcal{L}\left(\phi\mid D_{labeled}\right)=\sum_{k=1}^{C}\sum_{i=1}^{N}\log\mathcal{T}\left(\boldsymbol{X}_{k,i}\mid\boldsymbol{\eta}_k,\frac{\lambda_k+1}{\lambda_k\left(\nu_k-d+1\right)}\Psi_k,\nu_k-d+1\right),\tag{46}$$

Here, to generate a valid **Normal-inverse-Wishart (NIW) distribution**, $\lambda$ must be greater than 0, $\nu$ must be greater than $d-1$, and $\Psi$ must be positive definite.

### *3.3.2. Meta-Learning for Bayesian Quadratic Classifier*

The Bayesian quadratic classifier integrates with the feature extraction network trained via meta self-supervised learning. Using a cross-task meta-learning framework, the classifier extracts meta-knowledge through fragmented meta-learning to minimize expected risk, as defined below.

$$\phi^*=\arg\min_{\phi}\mathop{\mathbb{E}}_{\mathrm{Q}\sim p(X)}\left[-\sum_{k=1}^{C}\sum_{i=1}^{N}\log\mathcal{T}\left(\boldsymbol{X}_{k,i}\mid\boldsymbol{\eta}_k,\frac{\lambda_k+1}{\lambda_k\left(\nu_k-d+1\right)}\Psi_k,\nu_k-d+1\right)\right],\tag{47}$$

Here, Q represents the test set (query set) of each subtask drawn from the diagnostic task distribution.

## 3.4. Sample Selection Framework Based on Uncertainty Measurement

To effectively address the issue of Out-of-Domain (OOD) samples in fault diagnosis under data imbalance conditions, this paper proposes a sample filtering and model calibration framework based on distribution uncertainty quantification. The framework leverages Bayesian methods and Dirichlet Prior Networks to model the distribution uncertainty of generated samples (as discussed in Section 3.1) and the enhanced sample feature representations (as discussed in Section 3.2). This quantification determines whether samples belong to the training data distribution, providing a basis for sample filtering and further optimizing model performance (as discussed in Section 3.3). The detailed process is illustrated in Figure 8.

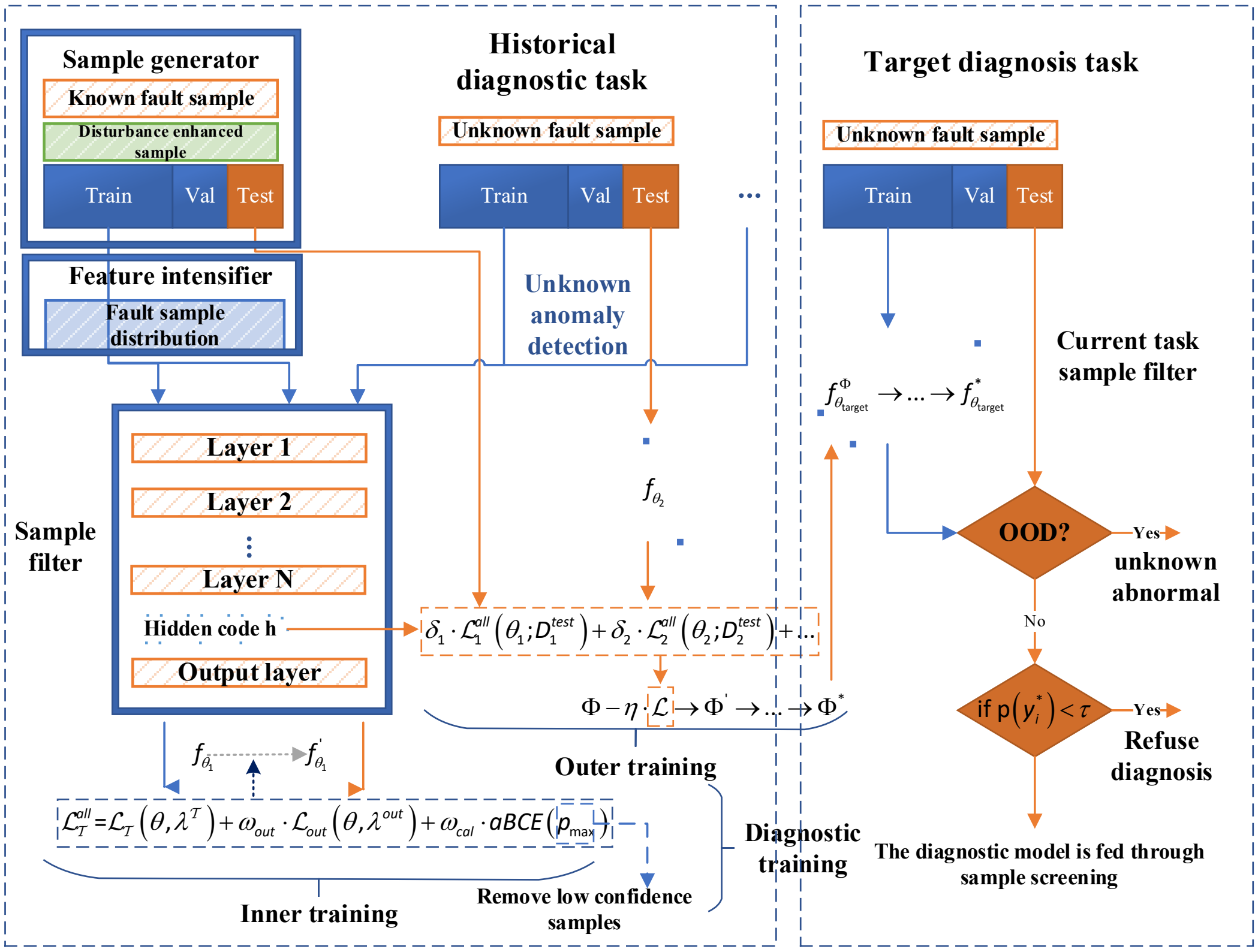

**Fig. 8.** Sample Filter Application Flow

The sample filtering framework serves as a preprocessing module for diagnostic models, ensuring high-quality data input for model calibration and diagnostic tasks. It operates in two stages: first, the OOD Detection Stage quantifies the distribution uncertainty of input samples, flagging high-uncertainty samples as Out-of-Domain (OOD) due to distribution shifts or unseen classes. These OOD samples are either removed or analyzed by experts to maintain data quality. Second, in the Low-Confidence Sample Removal Stage, non-OOD samples are evaluated for model output confidence. Samples with confidence scores below a threshold are removed to avoid overfitting or optimization errors during calibration. The framework integrates multiple synergistic steps: diagnostic inner training to optimize basic classification capabilities; OOD detection inner training to identify and remove OOD samples; calibration error inner training to adjust model confidence through adaptive calibration; domain-aware outer training to enhance generalization across new tasks and distributions; and sample filtering rules to ensure data reliability through uncertainty quantification and confidence evaluation. This cohesive process effectively addresses data imbalance and OOD challenges, significantly improving diagnostic accuracy and robustness.

#### 3.4.1. Inner-Loop Training for Diagnostic Tasks

The predictive function $f_{\theta}(x)$ serves as the core of the diagnostic task's inner-loop training, generating the probability distribution $p(y|x,\theta)$ for input samples. The training objective is formulated using the Reverse KL-divergence Loss (RKL) to optimize classification performance and capture distributional uncertainty. The loss is defined as:

$$\mathcal{L}^{RKL}(y,x,\theta;\alpha) = \sum_{c=1}^{C} I(y=c) \cdot KL\left[p(\mu|x;\theta) \parallel \text{Dir}\left(\mu|\alpha^{c}\right)\right], \tag{48}$$

where $\text{Dir}\left(\mu|\alpha^{c}\right)$ is the Dirichlet prior distribution for class $c_1$ and $\alpha^{c}$ is defined as:

$$\alpha_k^c = \begin{cases} \alpha + 1 & \text{if } c = k \\ 1 & \text{if } c \neq k \end{cases} \tag{49}$$

This formula measures the KL divergence between the predictive distribution $p(\mu|x;\theta)$ generated by $f_{\theta}(x)$ and the prior distribution, optimizing classification performance for indistribution samples.

To further enhance the model's capability to detect out-of-distribution (OOD) samples, the loss function is extended as:

$$\mathcal{L}^{RKL}\left(\theta;\alpha^{T},\alpha^{\text{out}}\right) = \mathbb{E}_{D_{\text{in}}}\left[KL\left[p(\mu|x;\theta) \parallel \text{Dir}\left(\mu|\alpha^{T}\right)\right]\right] + \omega_{\text{out}} \cdot \mathbb{E}_{D_{\text{out}}}\left[KL\left[p(\mu|x;\theta) \parallel \text{Dir}\left(\mu|\alpha^{\text{out}}\right)\right]\right], \tag{50}$$

Where, $f_{\theta}(x)$ represents the classifier's predictive model; $D_{\text{in}}$ and $D_{\text{out}}$ refer to in-distribution and out-of-distribution samples, respectively; $\alpha^{T}$ and $\alpha^{\text{out}}$ denote the prior parameters for in-distribution and out-of-distribution samples; $\omega_{\text{out}}$ balances the contribution of OOD detection.

By optimizing this loss function, the classifier $f_{\theta}(x)$ improves its classification accuracy for in-distribution samples and enhances its ability to detect low-confidence and OOD samples, providing reliable initial parameters for subsequent model calibration and outer-loop training.

#### 3.4.2. Integrated Inner-Loop Training for Diagnosis and OOD Detection

The integrated inner-loop training process optimizes the functions $f_{\theta_1}(x)$ for in-distribution classification and $f'_{\theta_1}(x)$ for out-of-distribution (OOD) detection, while incorporating calibration mechanisms. The total loss is defined as:

$$\mathcal{L}_T^{\text{all}} = \mathcal{L}_T\left(\theta,\lambda^{T}\right) + \omega_{\text{out}} \cdot \mathcal{L}_{\text{out}}\left(\theta,\lambda^{\text{out}}\right) + \omega_{\text{cal}} \cdot \text{aBCE}\left(p_{\max}\right) \tag{51}$$

where:

1. In-distribution Training Loss $\left(\mathcal{L}_T\left(\theta,\lambda^{T}\right)\right)$:

$$\mathcal{L}_T\left(\theta,\lambda^{T}\right) = \mathbb{E}_{D_{\text{in}}}\left[-\log p(y|x,\theta) - \frac{\lambda^{T}}{C}\sum_{c} \text{sigmoid}\left(z_c(x)\right)\right]. \tag{52}$$

This loss function optimizes the predictive function $f_{\theta_1}(x)$ to maximize classification accuracy while discouraging overconfidence using a regularization term weighted by $\lambda^{T}$.

2. OOD Detection Loss $\left(\mathcal{L}_{\text{out}}\left(\theta,\lambda^{\text{out}}\right)\right)$:

$$\mathcal{L}_{\text{out}}\left(\theta,\lambda^{\text{out}}\right) = \mathbb{E}_{D_{\text{out}}}\left[H_c(U,p(y|x,\theta)) - \frac{\lambda^{\text{out}}}{C}\sum_{c} \text{sigmoid}\left(z_c(x)\right)\right]. \tag{53}$$

This loss function trains the OOD detection function $f'_{\theta_1}(x)$ by encouraging uniform predictions for OOD samples (minimizing confidence), guided by $\lambda^{\text{out}}$.

3. Adaptive Calibration Loss $\left(\text{aBCE}\left(p_{\max}\right)\right)$:

$$\text{aBCE}\left(p_{\max}\right) = -\log\left(1 - p_{\max}\right) \tag{54}$$

which penalizes high-confidence predictions for uncertain samples to improve calibration reliability.

Parameter Update for $f'_{\theta_1}(x)$:

The predictive function $f'_{\theta_1}(x)$ is updated iteratively based on the gradient of the combined loss:

$$\theta'_1 = \theta_1 - \eta \nabla_{\theta_1} \mathcal{L}_T^{\text{all}} \tag{55}$$

where $\eta$ is the learning rate. This update process ensures that the parameters $\theta'_1$ adapt not only to the in-distribution classification task but also to OOD detection and calibration requirements. Through this mechanism, $f'_{\theta_1}(x)$ evolves to robustly identify and handle OOD samples while maintaining accurate and calibrated predictions for in-distribution data.

The final goal is to minimize the total loss:

$$\theta^* = \arg\min_{\theta} \mathcal{L}_T^{\text{all}} \tag{56}$$

yielding an optimized model that addresses in-distribution accuracy, OOD robustness, and prediction calibration simultaneously.

### 3.4.3. Domain-Aware Outer Training

To enhance the model's generalization ability across tasks and data domains, the outer training phase employs a domain-aware mechanism to construct a Bayesian feature alignment strategy for cross-task optimization. During this phase, the sample filter $\mathcal{F}_\theta$ is used to process the target task's training set $D_{\text{target}}$ and validation set $D_{\text{val},}$, ensuring high-quality samples are selected for outer training.

First, a global posterior distribution $p_{\text{global}}(\Phi)$ is introduced to capture shared features across different tasks. The posterior distribution is represented using the pairwise task similarity matrix $C_{ij}$, calculated as:

$$C_{ij} = \frac{h_i^{\top} h_j}{\| h_i \| \cdot \| h_j \|}, \quad \| C_T \|_F = \sqrt{\sum_{j=1}^{K} C_{ij}^2} \tag{57}$$

where $\| C_T \|_F$ is the Frobenius norm of task similarity, measuring the dependency between tasks.

Next, the task similarities are normalized using the softmax function to derive task importance weights $\delta_T$ :

$$\delta_T = \frac{\exp(-\| C_T \|_F / T)}{\sum_{j=1}^{K} \exp(-\| C_j \|_F / T)} \tag{58}$$

Here, $T$ is a temperature parameter that adjusts the distribution of weights across tasks.

Finally, task losses are aggregated using these weights to optimize the global posterior distribution $p_{\text{global}}(\Phi)$ and update the model initialization parameter $\Phi$ :

$$\Phi \leftarrow \Phi - \eta \sum_{T=1}^{K} \delta_T \nabla_\Phi \mathcal{L}_{\text{out}}(\Phi, D_{\text{target}}) \tag{59}$$

where $\mathcal{L}_{\text{out}}$ includes the target task loss and a KL divergence term for task alignment:

$$\mathcal{L}_{\text{out}} = \mathcal{L}_{\text{task}} + \lambda \cdot \text{KL}\left(p(h_T|\Phi) \,\|\, p_{\text{global}}(\Phi)\right) \tag{60}$$

Through the domain-aware outer training process, the model learns to extract shared features across tasks while optimizing for specific tasks, enabling rapid adaptation to new tasks and diverse data distributions. This significantly improves the model's robustness and generalization capability in various scenarios.

### 3.4.4. Sample Filtering Rules

After completing the cross-task outer training, the initialized sample filter $\mathcal{F}_\theta$ is employed to filter the training set $D_{\text{train}}^{\text{target}}$ and validation set $D_{\text{val}}^{\text{target}}$ for the target task, ensuring high-quality samples for final training.For unseen samples in the validation set, $\mathcal{F}_\theta$ first conducts Out-of-Domain (OOD) detection by calculating the sample's predictive maximum probability $p_{\max}$. This is used to determine whether the sample belongs to the known domain. If the sample is classified as OOD, it is rejected as:

$$\text{output}\left(x_i\right)=\begin{cases}x_i, & \text{if } p\left(y_i\right)\geq\tau\\ \text{reject}, & \text{otherwise}\end{cases} \tag{61}$$

where $\tau\in[0,1]$ is the confidence threshold.

If a sample is classified within the domain but fails to meet the confidence threshold $\tau$, it is further calibrated using the maximum confidence $p_{\max}$. This ensures that the diagnosis model operates only on samples with a confidence level above the threshold $\tau$. Samples below this threshold are rejected to maintain the accuracy and reliability of the diagnostic task.

Ultimately, the confidence rate achieved by the diagnostic model on the validation set serves as an indicator of the model's adequacy. If the confidence level on the validation set meets the filtering criteria, the diagnostic model is considered effective.

# 4. Case Setup

This study uses a series of cases to systematically validate the stability, generalization capability, and interpretability of the proposed model in small-sample cross-task fault diagnosis. The cases include a primary case, an ablation case, a parameter sensitivity analysis case, and a visualization case.The **primary case** involves executing Any-way 1-5-shot evaluations on ten different datasets. This case verifies the generalization capability of the UBMF method in small-sample diagnostic scenarios, demonstrating its effectiveness in handling data imbalance and sample scarcity issues.The **ablation case** involves removing different modules from the model to observe their impact on overall performance, thereby verifying the effectiveness of each module.The **parameter sensitivity analysis case** involves gradually increasing the number of shots from 1 to 5. This demonstrates the model's stability and reliability when handling small-sample tasks and imbalanced data.Finally, the **visualization case** extracts features from the perturbed data and uses PCA for dimensionality reduction. This case shows that data perturbation enhances the contrastive learning effect, increases the feature differences between different operating conditions, and improves the generalization capability of the model. It also demonstrates the model's ability to capture weak signals, thereby enhancing interpretability.

## *4.1. Experimental Design*

This paper addresses the challenge of cross-device and cross-object variable operating condition data imbalance diagnosis by proposing a comprehensive small-sample diagnostic method under data imbalance, referred to as **UBMF**. This method is designed for typical **cross-task** scenarios, requiring the model to learn knowledge from known devices or operating conditions and transfer it to different devices, conditions, or unknown fault modes. UBMF extracts robust and transferable features through cross-task self-supervised learning, enhances classification capabilities using a Bayesian quadratic classifier, and corrects data distribution biases while quantifying uncertainty through a sample filtering mechanism, thereby significantly improving diagnostic reliability and generalization. The complete implementation process and pseudocode of the method are detailed in **Appendix 1**, which provides a step-by-step description from feature extraction to classifier modeling and sample filtering, offering actionable details to guide engineering implementation. The specific settings of the 10 datasets used in this study are presented in **Appendix 2**, covering typical fault patterns across diverse industrial scenarios with rich variations in operating conditions and task differences. The cross-task diagnostic scenarios, as shown in **Table 1**, simulate the transferability from known tasks to unknown tasks, including device-level transfer, object-level transfer, and condition-level transfer, as well as diagnostic capability transfer for unknown samples. This experimental design validates the adaptability of the UBMF method under data imbalance conditions and demonstrates its generalization performance in complex industrial environments.

**Table 1**
Data Imbalance Diagnosis Task Settings

| No. | Task Design/Phase | Description |
|---|---|---|
| 1. Task Design | 1.1 Segmentbased Training | According to the "Any-way-1-5-shot" sampling rule, samples are extracted. Each task constructs a small support set from a few samples, simulating diverse task scenarios. |
| | 1.2 Batch Training | A batch of small sample sets is directly extracted from the dataset to construct multiple tasks, training the model to adapt to learning difficulties caused by imbalanced data. |
| 2. Test Task Construction | 2.1 Sampling from Meta-Test Dataset $D_{\text{meta - test}}$ | From the meta-test dataset $D_{\text{meta - test,}}$ $\mathbf{100}$ test tasks $T_i$ are sampled. Each query set $Q$ consists of $\mathbf{100}$ samples. Steps: First, randomly select the number of categories $N \in$ [2,MaxN], representing the number of classes for the task. Then, select $N$ categories for subsequent task sampling. |
| | 2.2 Constructing Support Set $S_i$ | The number of samples is imbalanced: some categories have abundant samples, while others are sparse. When constructing the support set $S_i$, the overall imbalance is preserved. For fault category health states, randomly select $K_i$ $\in [1,5]$, representing $K_i$ samples per category, forming the set $S_i$. |
| | 2.3 Constructing Query Set $Q_i$ | The query set $Q_i$ is constructed by randomly selecting $\mathbf{50}$ samples from balanced fault state categories. |
| 3. Model Testing Phase | 3.1 Task Execution | Use the support set $S_i$ for meta-learning (simulating fewshot learning under imbalanced data conditions). Evaluate model performance using the query set $Q_i$ and compute the standardized accuracy. |
| | 3.2 Performance Evaluation | Compute the average accuracy of $\mathbf{100}$ test tasks $T_i$ and calculate the $\mathbf{95\%}$ confidence interval as the model performance evaluation indicator. |
| 4. Scenario Verification for Diagnosis | 4.1 Downstream Diagnosis Task Verification (MainCase) | Select one dataset as the meta-test set and its subset as the training set, generating $\mathbf{10}$ types of task scenarios. Test the model's diagnostic performance under specific task backgrounds. |
| | 4.2 Unknown Sample Diagnosis (Visualization Analysis) | To verify the model's performance on unseen fault categories, $\mathbf{11}$ base fault categories are used as the test set, ensuring the model does not encounter these fault types during training. The dataset is divided into: Train, Valid, and Test, where the fault types in $S_1$ do not ap ${}^{r}\!\downarrow^{r}$ in $S_2$ or $S_3$. Visualization tools analyze the model's generalization performance on unknown categories. |

## *4.2. Model Design*

To implement the proposed method and validate its experimental effectiveness, the feature extraction backbone network, metric network, and sample selector were instantiated as three non-diagnostic task models. Beyond the Bayesian-based fault diagnosis model, these components were designed with specific network structures, while allowing flexibility for replacement with alternative architectures to accommodate different experimental needs.

(1) Backbone Network for Feature Extraction

This chapter employs a Wide Residual Network (WRN), specifically the WRN12 model with a width factor of 10, as the backbone network. Compared to ResNet, WRN increases the number of filters per residual block, enhancing feature extraction and representation without significantly increasing depth, ensuring computational efficiency. Trained with a meta self-supervised approach, the network effectively leverages unlabeled and task-domain data, further improving adaptability and representational capacity.

(2) Metric Network Model for Feature Enhancement

For the metric network, a simple combination of convolutional layers and fully connected layers is employed, as shown in **Error! Reference source not found.**.

**Table 2**

Metric Encoder Network Architecture

| Layer Number | Hyperparameters |
|---|---|
| 1 | Conv1d(INPUT，32，k=3，p=1)；BatchNorm1d()；<br>MaxPool1d(k=2，s=2)；LeakyReLU(0.2) |
| 2 | Dense(128)；BatchNorm1d()；ReLU()； |
| 3 | Dense(1)；Sigmoid() |

(3) Sample Selector

The sample selector captures domain distribution differences, enhancing the use of similar datasets across diagnostic tasks. This reduces reliance on small-sample optimization, shifting focus to outer-layer training while minimizing inner-layer adjustments. A deeper VGG network is used for feature extraction, effectively perceiving task-domain structures, as shown in Table 3.

**Table 3**

VGG Network Architecture Used in the UBML Sample Filtering Model

| layers | Hyperparameters |
|---|---|
| 1 | Conv2d(64,k=(3,3),p=same);ReLU();BatchNorm2d();Dropout(0.3) |
| 2 | Conv2d(64,k=(3,3),p=same);ReLU();BatchNorm2d();MaxPool2d(p=(2,2)) |
| 3 | Conv2d(128,k=(3,3),p=same);ReLU();BatchNorm2d();Dropout(0.4) |
| 4 | Conv2d(128,k=(3,3),p=same);ReLU();BatchNorm2d();MaxPool2d(p=(2,2)) |
| 5 | Flatten() |
| 6 | Dense(128);ReLU();BatchNorm2d();Dropout(0.5) |
| 7 | Dense(num of class); |

## *4.3. Comparative Baseline*

To demonstrate the performance of the proposed UBMF method, the following methods are used as baselines for comparison:

**Table 4**

Baselines

| | | | Main Case | Ablation Study | Visualization Analysis |
|---|---|---|---|---|---|
| 1 | MatchingNet | Matching networks with the same backbone network. The classification layer uses a two-layer fully connected network (Vinyals et al., 2016). | √ | | |
| 2 | ProtoNet | Prototypical networks with the same backbone network. The classification layer uses a two-layer fully connected network (Snell et al., 2017). | √ | | |
| 3 | MAML | (AE) with unsupervised reconstruction pre-training using the same backbone network, followed by supervised meta-learning using MAML (Sung et al., 2018). | √ | | |
| 4 | UBMF-E | UBMF model without the metric encoder, using Euclidean distance directly for pseudo-label propagation. | | √ | |
| 5 | UBMF-P | UBMF model without the pseudo-supervised loss from perturbation injection. | | √ | |
| 6 | UBMF-L | UBMF model using Linear Discriminant Analysis (LDA) as the classifier. | | √ | |

| | | | Main Case | Ablation Study | Visualization Analysis |
|---|---|---|---|---|---|
| 7 | UBMF-M | UBMF model using Quadratic Discriminant Analysis (QDA) without prior inference. | | √ | |
| 8 | BCnet | A binary classification model using the same backbone network, trained solely for anomaly detection. | | | √ |
| 9 | MaxP | A sample selector that determines OOD samples based solely on maximum probability. | | | √ |
| 10 | MI | A sample selector that determines OOD samples based solely on mutual information. | | | √ |
| 11 | DM | A diagnostic network that determines OOD samples based on maximum probability . | | | √ |
| 12 | DE | A sample selector that uses both differential entropy and maximum probability, requiring both to fall below a threshold to determine OOD samples. | | | √ |

# 5. Results Analysis

## 5.1. *Main Case Results*

The Any-way 1-5-shot results of each method on the ten datasets are shown in Figure 9, with detailed results available in Appendix E. It can be observed that the proposed UBMF method achieved the best results across all datasets.

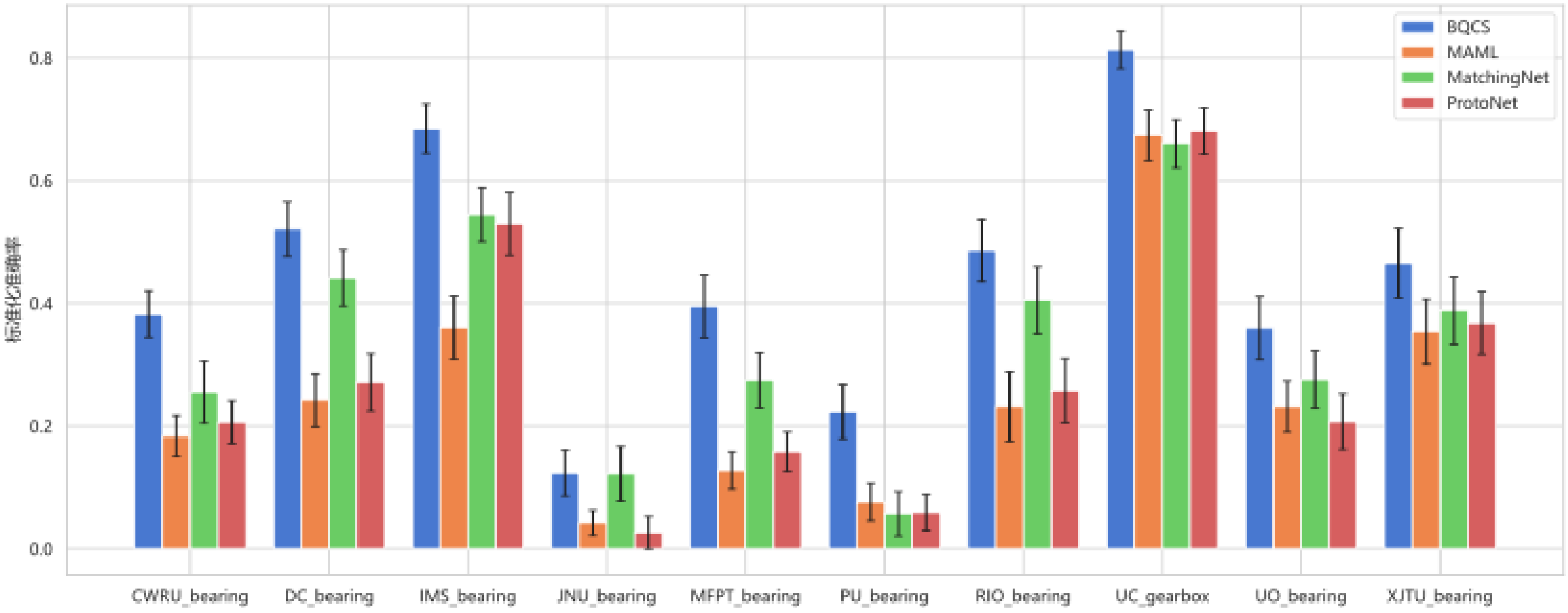

Fig. 9. Any-way 1-5-shot Case Results

The comparison of the improvement in standardized accuracy of the UBMF method relative to the best-performing baseline method is shown in **Error! Reference source not found.**. It can be seen that in the vast majority of scenarios, the UBMF method demonstrates significant advantages, achieving an average improvement of 42.22% across ten Any-way 1-5-shot small-sample diagnostic scenarios.

**Table 5**
Standardized Accuracy Improvement (%) over Baseline Methods on Each Dataset

| Dataset | Improvement | Dataset | Improvement |
|---|---|---|---|
| XJTU_bearing | 19.92 | MFPT_bearing | 43.90 |
| UO_bearing | 30.49 | JNU_bearing | 1.23 |
| UC_gearbox | 19.47 | IMS_bearing | 25.72 |
| RIO_bearing | 20.09 | DC_bearing | 18.11 |
| PU_bearing | 193.42 | CWRU_bearing | 49.88 |
| Average | | 42.22 | |

## 5.2. *Ablation Study*

To further verify the effectiveness of each module of the UBMF method, the settings in table 4 are used for the ablation study: The results of the ablation study are shown in Figure 10. It can be observed that removing any of the proposed modules affects the performance of the UBMF method. Across all datasets, the ablation did not achieve consistent results, and removing certain modules even led to improvements in specific datasets. This is mainly due to the inherent randomness in model training under small-sample conditions. A further summary of the ranking of each method is presented in Table 6. It can be seen that the proposed method is the most optimal in terms of generalization and achieves significant overall superiority, demonstrating the effectiveness of each module and the advantage of integrating them into the UBMF method.

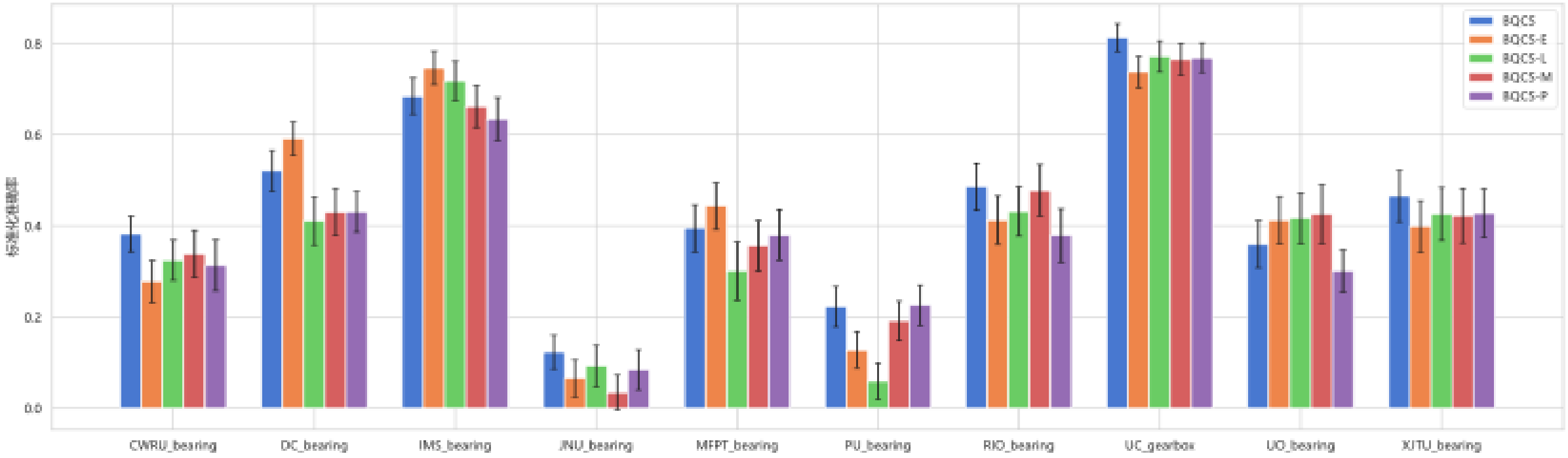

**Fig. 10.** UBMF Ablation Study Results

**Table 6**
Average Ranking of Each Method in Ablation Study

| Method | UBMF | UBMF-L | UBMF-E | UBMF-M | UBMF-P |
|---|---|---|---|---|---|
| Ranking | 1.8 | 3.2 | 3.3 | 3.3 | 3.4 |

## 5.3. *Parameter Sensitivity Analysis*

Analysis of Different Way and Shot Results To further analyze the results for different shot conditions, a balanced small-sample diagnostic case was conducted using a fixed Max N-way K-shot approach. The evaluation used the general diagnostic accuracy metric, as shown in Figure 11 to Figure 15. It can be observed that, in the 1-shot condition, there is considerable randomness in model training, and the differences between various ablation methods are not obvious. However, as the number of shots increases, the advantages of the complete UBMF method gradually become apparent.

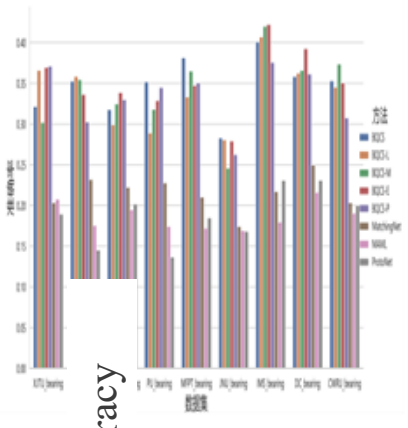

Accuracy

Method

Data

**Fig. 11.** 1-shot Case Results

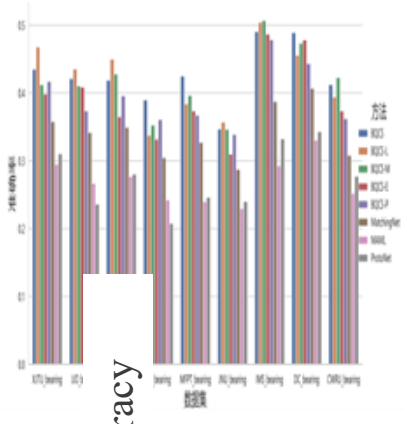

Accuracy

Method

Data

**Fig. 12.** 2-shot Case Results

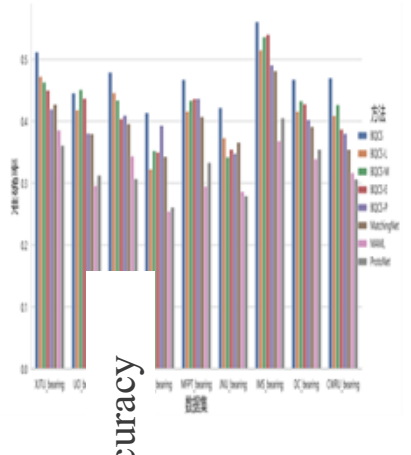

Accuracy

Method

Data

**Fig. 13.** 3-shot Case Results

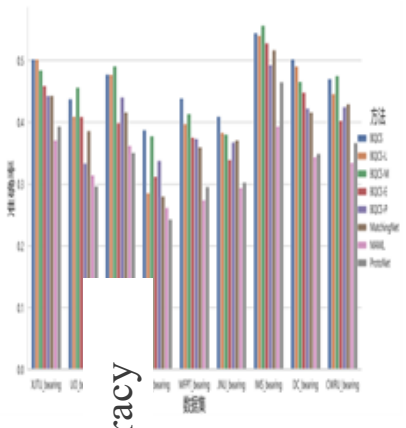

**Fig. 14.** 4-shot Case Results

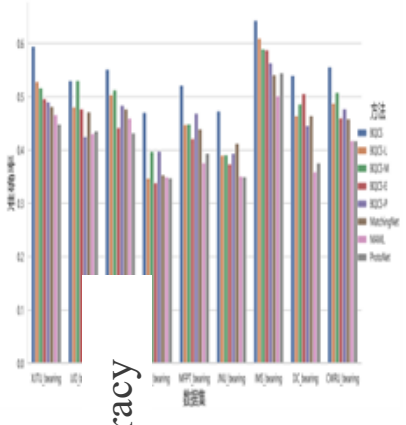

**Fig. 15.** 5-shot Case Results

Taking the MFPT_bearing task as an example, the specific diagnostic results are analyzed. The distribution of standardized diagnostic accuracy for 100 meta-testing tasks is shown in Figure 16. Some of the testing results have standardized accuracies below 0 (all of which are from the 1-shot condition), indicating that the training samples extracted for these meta-testing tasks do not well represent the overall distribution. The significant variation in the distribution of overall accuracy indicates that under small-sample conditions, data-driven diagnostic inference has inherent randomness that is difficult to eliminate.

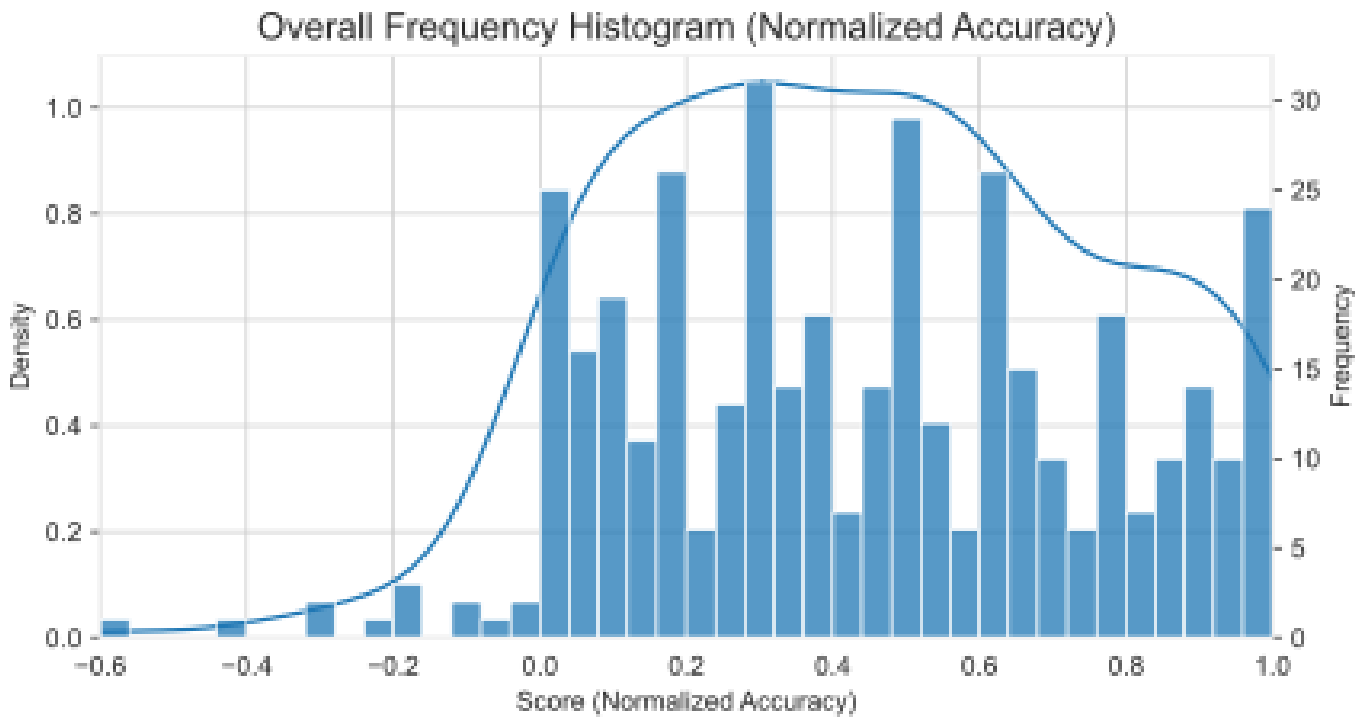

**Fig. 16.** Distribution of Standardized Accuracy Results for 100 Meta-Testing Runs on the MFPT_bearing Task

The calibration error for 100 meta-testing tasks is shown in Figure 17. It can be seen that under the 1-5-shot conditions, the uncertainty calibration error is relatively high, indicating that the model struggles to predict future generalization capabilities from a very small number of labeled samples.

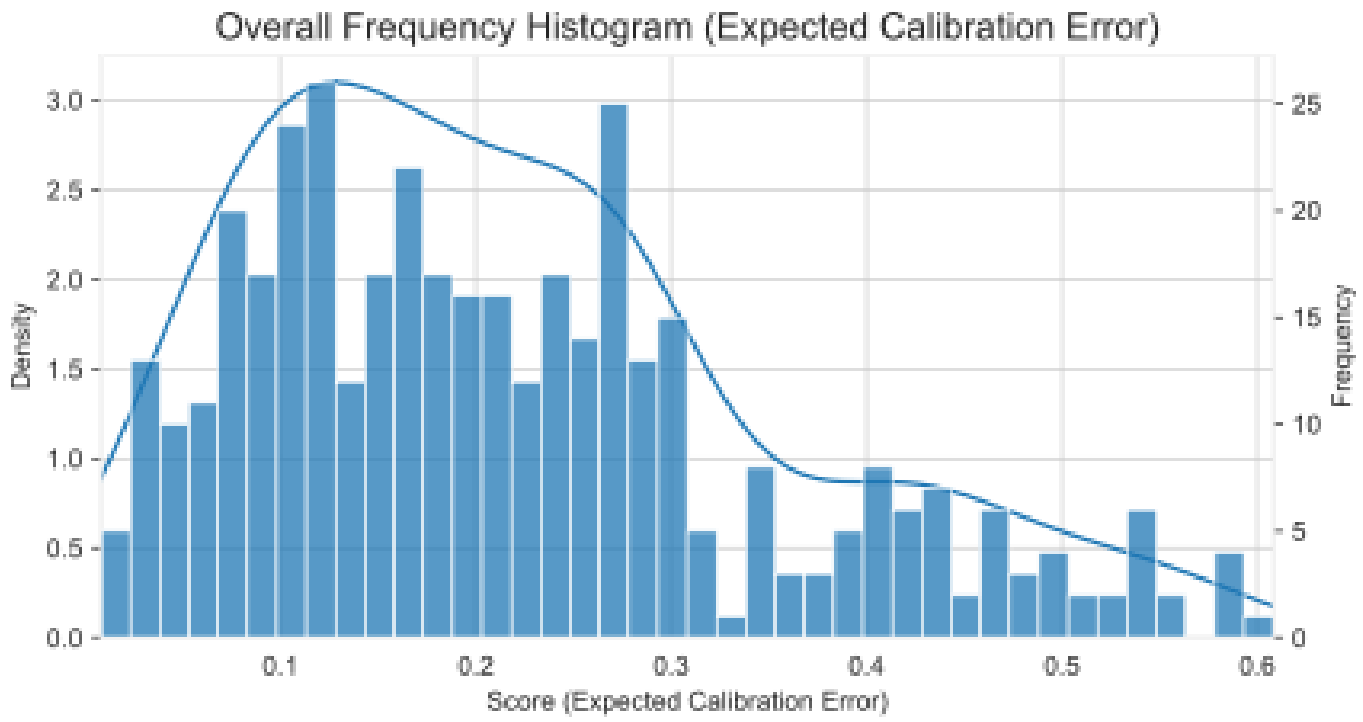

Fig. 17. Uncertainty Calibration Results Distribution

The distribution of diagnostic performance for different total numbers of healthy states (ways) is shown in Figure 18. It can be clearly observed that increasing the number of healthy state categories significantly increases the difficulty of diagnosis. Furthermore, using diagnostic accuracy alone as a metric can lead to a distorted evaluation of performance.

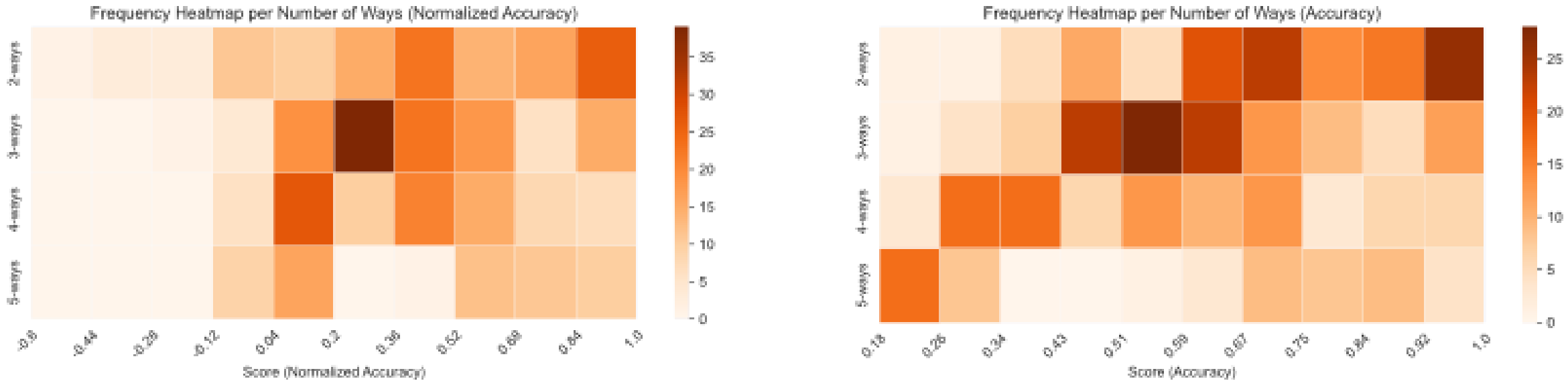

**Fig. 18.** Distribution of Standardized Diagnostic Accuracy (left) and Diagnostic Accuracy (right) for Different Numbers of Classes (Ways)

The distribution of diagnostic performance for different numbers of samples per class (shots) is shown in Figure 19. It can be clearly observed that increasing the data volume improves the overall diagnostic performance. However, significant randomness still exists, and the diagnostic model cannot guarantee that a single increase in data will necessarily lead to performance improvement.

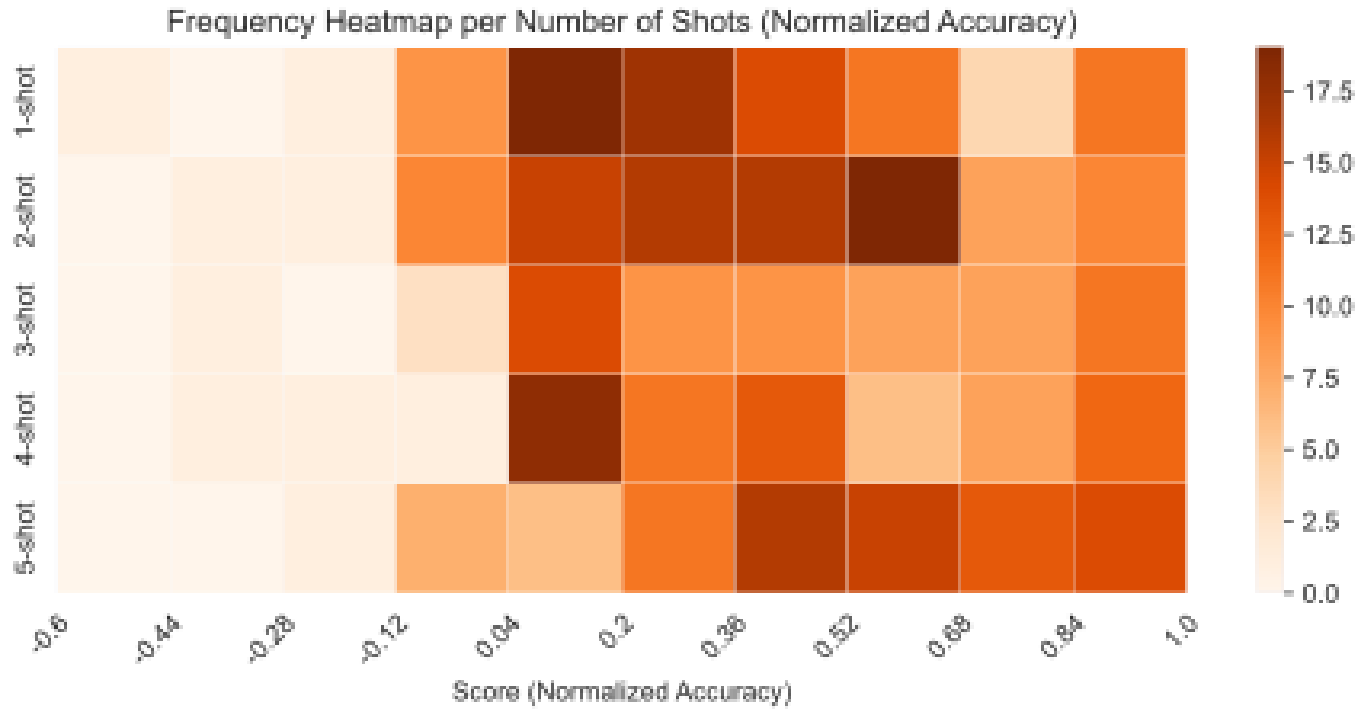

**Fig. 19.** Distribution of Results for Different Numbers of Samples per Class (Shots)

## 5.4. *Visualization Analysis*

### 5.4.1. *Feature Enhancement*

During the training process of contrastive learning, this study injected different types of perturbation results into the cross-task self-supervised feature extraction process. Specifically, the strong and weak perturbation methods previously described were used to perturb the original data, and the perturbed data were generated as new training samples and merged into the training set. To verify the effect of data augmentation based on data perturbation on enhancing the contrastive learning outcome, we selected 100 fault data samples from the XJTU bearing dataset as a small batch of original samples. We preset different types of perturbations and gradually injected these perturbations at a rate ranging from 20% to 80%. Two faulty bearings were randomly selected as observation subjects (Sample 1 in Figure 20 and Sample 2 in Figure 21). Then, the 128-dimensional features extracted by the cross-task self-supervised feature extractor were reduced using PCA for dimensionality reduction and were visualized.

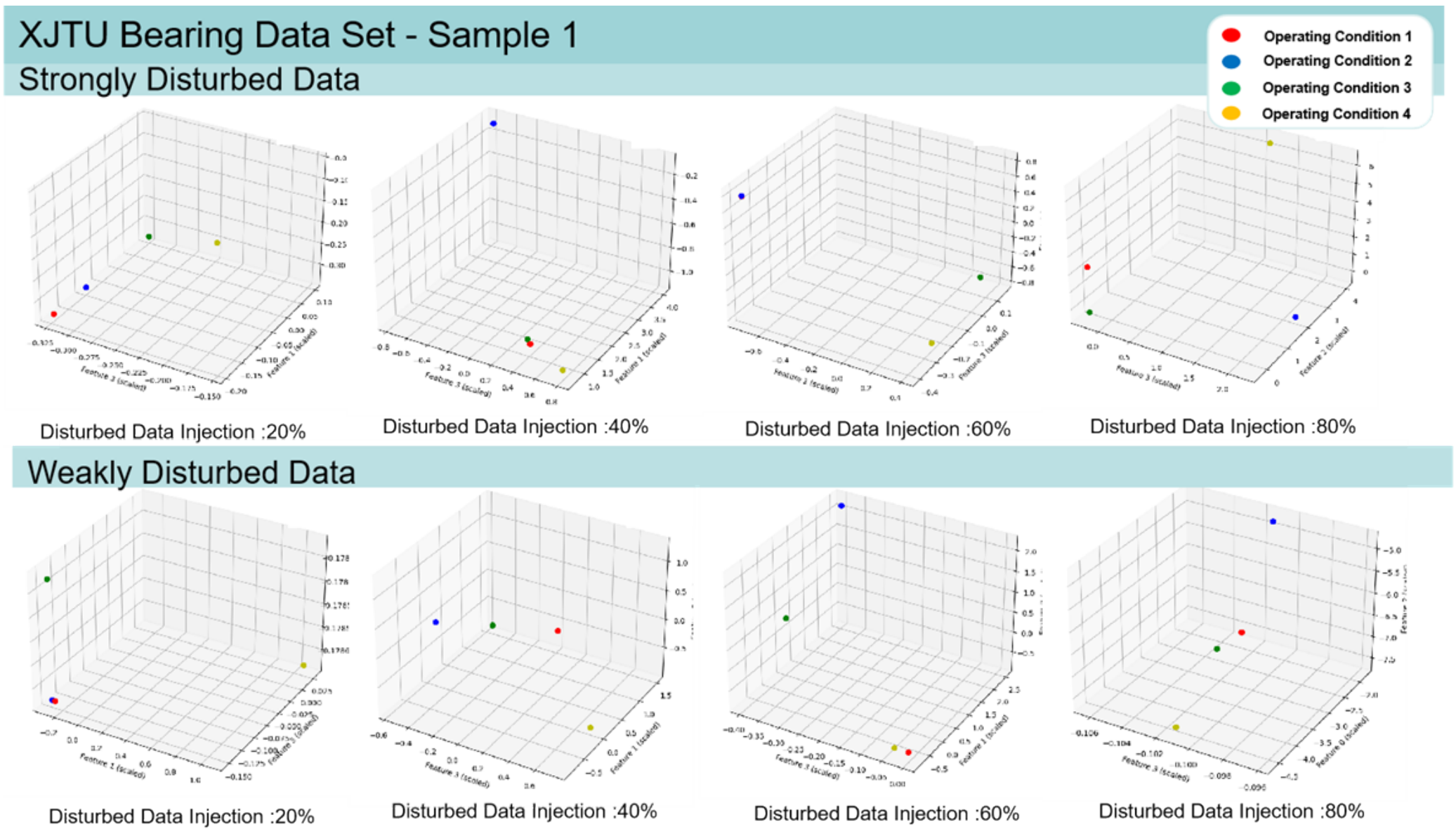

**Fig. 20.** The Feature Enhancement of XJTU Sample1

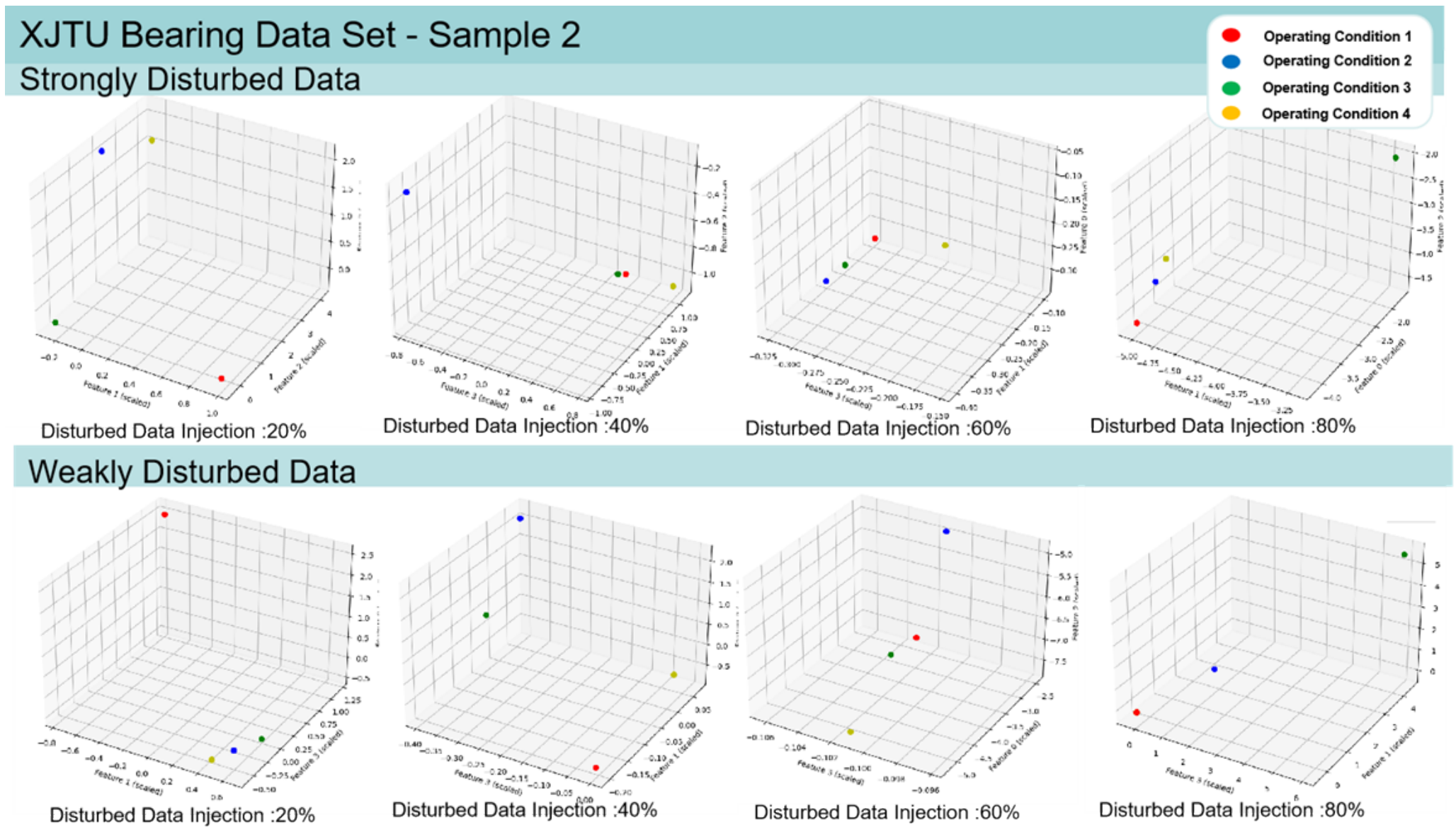

**Fig. 21.** the Feature Enhancement of XJTU Sample2

From the results in Figure 20 and Figure 21, it can be observed that for the same observation object under four different operating conditions, the relative distance between features of different fault classes in Euclidean space significantly increases with an increase in the perturbation injection ratio. This indicates that data augmentation through data perturbation can significantly enhance the effect of contrastive learning. The perturbation introduces more diverse fault features, increasing the feature differences between different fault classes. This helps the model distinguish different fault classes more effectively, enhancing the model's generalization capability and its ability to capture weak signals.

### *5.4.2. The Unknown Sample Diagnosis*

In the unknown sample diagnosis task, the primary goal is to evaluate whether the sample selector can identify anomalous samples in unseen tasks. The performance is assessed using the AUROC (Area Under the Receiver Operating Characteristic curve), treating anomalies as positive cases and known healthy states as negative cases. Considering the similarity of fault modes in bearing datasets, datasets with more fault modes were selected for testing to better validate the proposed method. The AUROC results across datasets are shown in Figure 22, with the average rankings of methods summarized in Table 16. The proposed method achieves the best performance on most datasets, with comparable results occasionally obtained by the binary classification model BCnet. This indicates that standalone anomaly detection is often easier than jointly optimizing anomaly detection and classification.

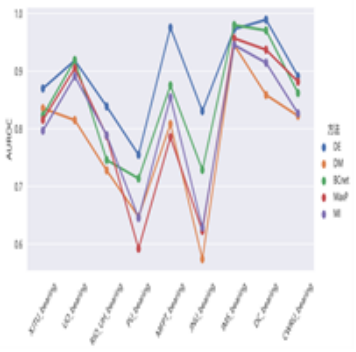

Method

**Fig. 22.** Major results of unknown anomaly detection

**Table 7**
AUROC Rankings of Unknown Anomaly Detection Methods

| baselines | DE | BCnet | MaxP | MI | DM |
|---|---|---|---|---|---|
| rank | 1.22 | 2.22 | 3.56 | 3.72 | 4.28 |

The performance of uncertainty metrics in measuring distribution uncertainty was analyzed, showing that incorporating differential entropy into the loss function, combined with maximum probability for joint evaluation, enhances the identification of unknown anomalies. For further analysis, the DC_bearing dataset was selected, treating normal, inner ring, and outer ring health states as known, while rolling element faults and four combination faults were considered unknown anomalies. Additional samples from other datasets were used as training data for unknown anomalies. Figure 23 visualizes the features of the trained sample classifier using t-SNE, demonstrating effective separation between known and unknown health states in the feature space.

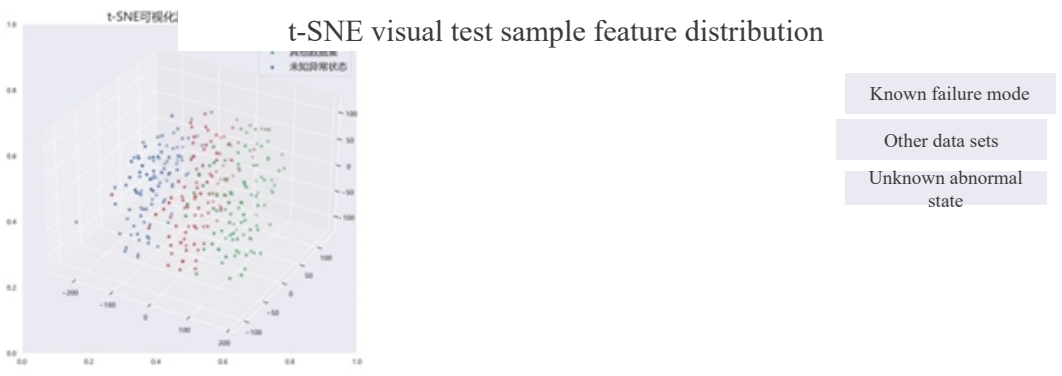

**Fig. 23.** t-SNE visualization of features extracted by sample filters

Figure 24 (top left) shows the feature distribution of known ("ID") and unknown ("OOD") samples, where distinct shapes represent different health states. The sample selector effectively distinguishes ID from OOD samples but struggles to differentiate specific health states. Heatmaps in Figure 24 reveal that differential entropy provides better separation between ID and OOD samples, validating the use of a Dirichlet distribution in the loss function. Figure 25 shows that while anomaly detection performance initially improves, it declines in later stages, reflecting a conflict between the optimization objectives of anomaly detection and diagnosis.

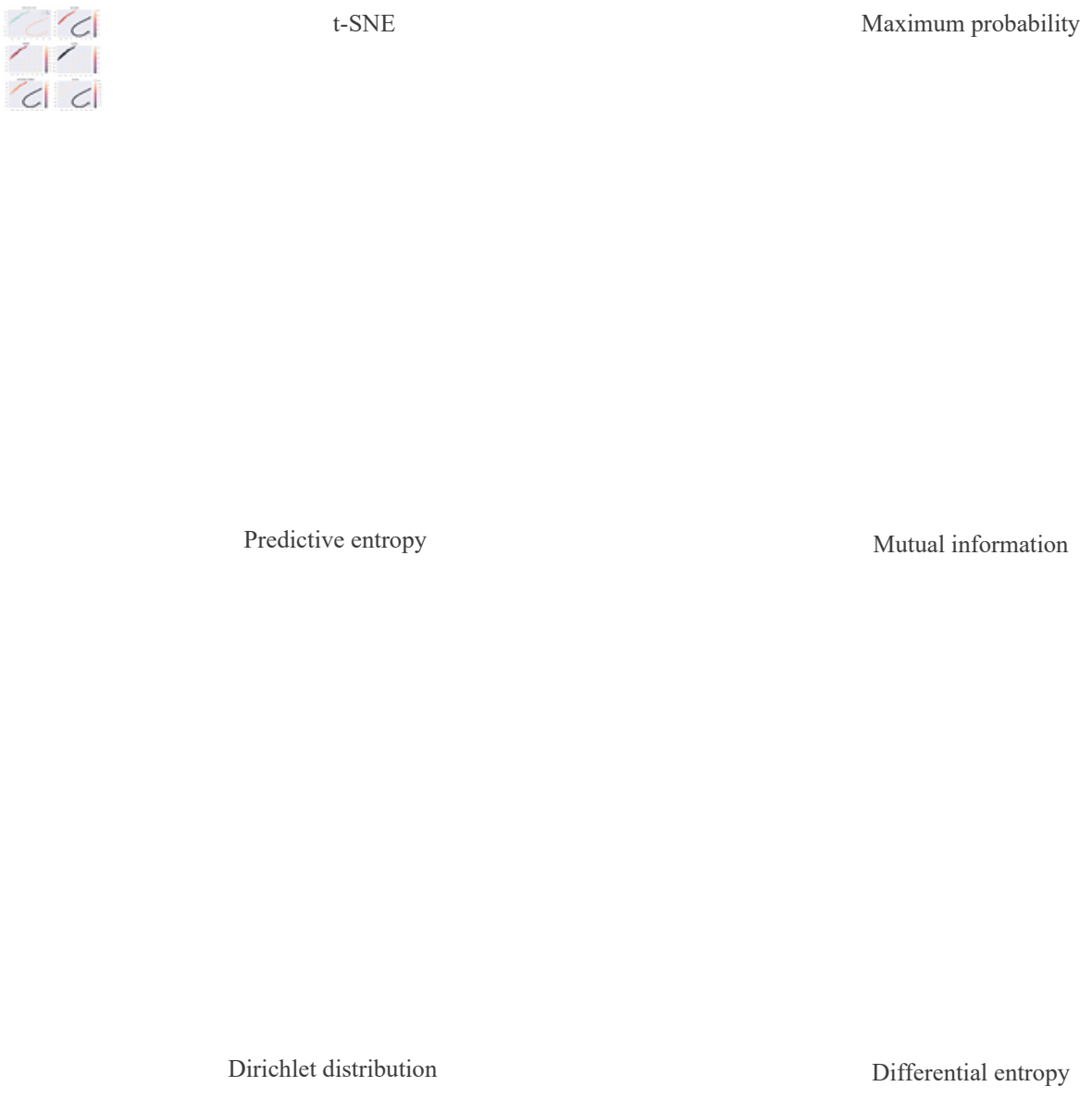

**Fig. 24.** Sample filter feature distribution and uncertainty measure for known and unknown samples

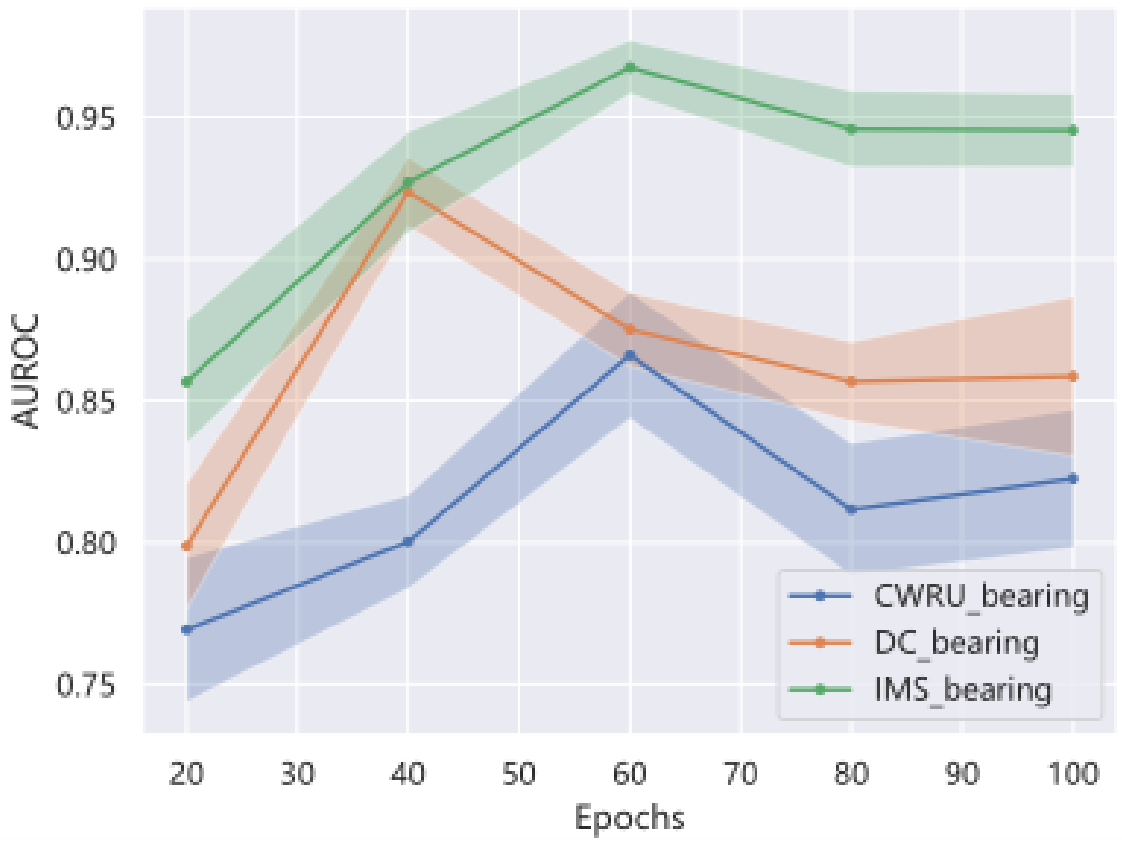

**Fig. 25.** The abnormal detection ability of each training stage of the diagnostic model

### *5.4.3. Rejection Diagnosis*

This section evaluates whether the sample selector can calibrate the model in unknown sample diagnosis tasks. With varying rejection thresholds, the trained sample selector filters test task data before passing it to the diagnostic model for evaluation. Results in Table 17 show that, for the five tasks with pre-selection accuracy below 70%, the proposed sample selector effectively ensures that post-threshold samples achieve or approach the target accuracy level.

**Table 8**

Diagnostic Results (Accuracy) of Filtered Samples

| Threshold | IMS_bearing | UO_bearing | JNU_bearing | PU_bearing | BUAA_gearbox |
|---|---|---|---|---|---|
| 0.7 | 74.48 | 75.20 | 74.20 | 68.00 | 73.42 |
| 0.8 | 83.63 | 82.85 | 83.30 | 83.70 | 80.53 |
| 0.9 | 87.73 | 92.90 | 93.55 | 90.15 | 93.66 |

To further examine the uncertainty calibration of the sample selector itself, the calibration was evaluated on the test set during training, as shown in Figure 26. The results indicate that incorporating calibration loss during training gradually shifts the distribution toward 45 degrees, representing better calibration.

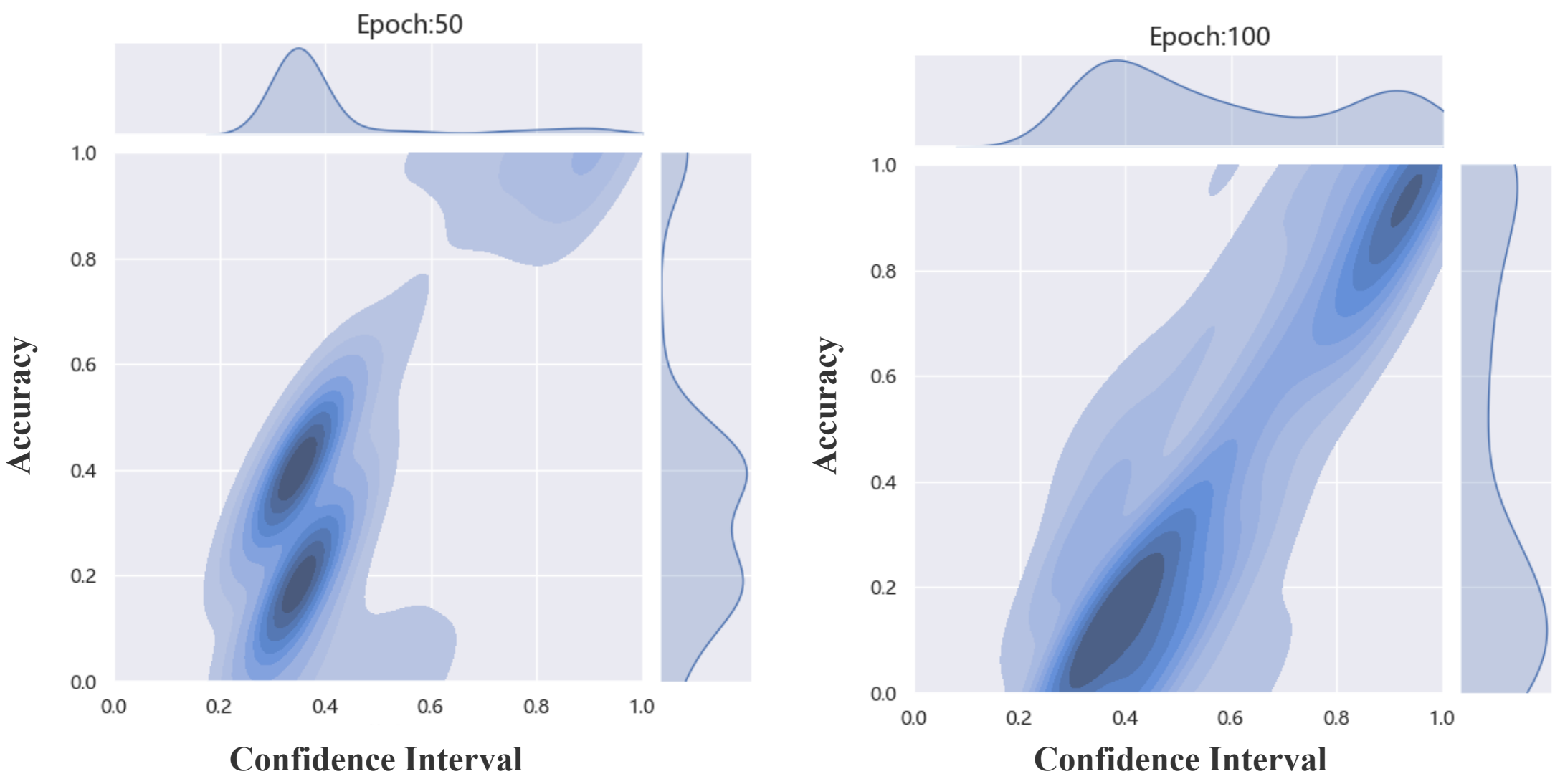

**Fig. 26.** Sample filter training procedure

When the calibration loss is removed from the inner-layer training of the sample selector, the training process changes as shown in Figure 27. Although the training progresses faster from the perspective of diagnostic accuracy, the gap between predicted confidence and actual diagnostic accuracy increases, indicating that the model cannot effectively evaluate its diagnostic capability.

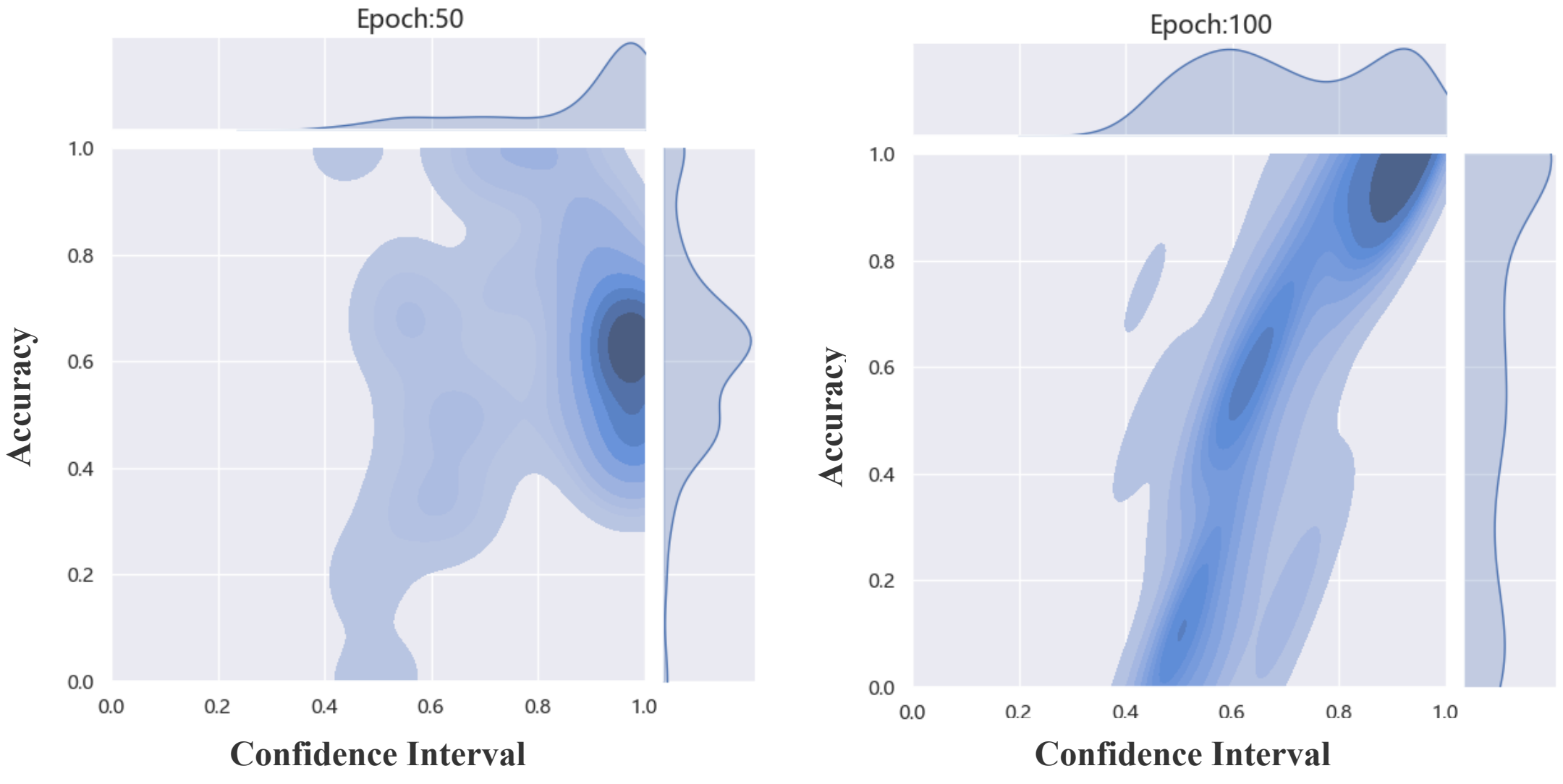

**Fig. 27.** Sample filter training process with calibration loss removed

# 6. Conclusion

The UBMF method offers a practical solution for fault diagnosis under data imbalance, particularly in industrial scenarios with limited and imbalanced data. Leveraging Bayesian meta-knowledge extraction, pseudo-label propagation, cross-task self-supervised learning, and uncertainty regularization, it enhances fault data augmentation, feature extraction, and diagnostic efficiency while reducing reliance on manual annotation. A key innovation is its uncertainty-based sample filtering, which prioritizes high-quality samples to mitigate data imbalance, improve feature learning, and enhance sensitivity to minority-class patterns. By integrating domain knowledge with imbalanced data using consistency regularization, UBMF ensures reliable, interpretable diagnostics and robust performance. Experimental results show a 42.22% improvement over baselines, demonstrating UBMF's scalability and adaptability to diverse operating conditions. It provides a cost-effective, transferable solution for multi-condition and multi-equipment fault diagnosis, offering significant industrial value by improving efficiency and reducing downtime.

# References

Azari, M. S., Santini, S., Edrisi, F., & Flammini, F. (2025). Self-adaptive fault diagnosis for unseen working conditions based on digital twins and domain generalization. *Reliability Engineering & System Safety*, *254*, 110560. https://doi.org/10.1016/j.ress.2024.110560

Bajaj, N. S., Patange, A. D., Jegadeeshwaran, R., Pardeshi, S. S., Kulkarni, K. A., & Ghatpande, R. S. (2023). Application of metaheuristic optimization based support vector machine for milling cutter health monitoring. *Intelligent Systems with Applications*, *18*, 200196. https://doi.org/10.1016/j.iswa.2023.200196

Benedict, M., Benedict, M., Moylan, S., Al-Abed, S., Ashforth, C., Chou, K.,…Lewis, A. (2022). *The strategy for American leadership in high-consequence additive manufacturing*. US Department of Commerce, National Institute of Standards and Technology. https://nvlpubs.nist.gov/nistpubs/ams/NIST.AMS.600-10.pdf

Cao, X., Peng, K., & Jiao, R. (2024). Degradation modeling and remaining life prediction for a multi-component system under triple uncertainties. *Computers & Industrial Engineering*, *195*, 110432. https://doi.org/10.1016/j.cie.2024.110432

Chan, J., Papaioannou, I., & Straub, D. (2024). Bayesian improved cross entropy method with categorical mixture models for network reliability assessment. *Reliability Engineering & System Safety*, *252*, 110432. https://doi.org/10.1016/j.ress.2024.110432

Che, Q.-H., Nguyen, L.-C., Luu, D.-T., & Nguyen, V.-T. (2025). Enhancing person re-identification via Uncertainty Feature Fusion Method and Auto-weighted Measure Combination. *Knowledge-Based Systems*, *307*, 112737. https://doi.org/10.1016/j.knosys.2024.112737

Chen, Z., Chen, J., Feng, Y., Liu, S., Zhang, T., Zhang, K., & Xiao, W. (2022). Imbalance fault diagnosis under long-tailed distribution: Challenges, solutions and prospects. *Knowledge-Based Systems*, *258*, 110008. https://doi.org/10.1016/j.knosys.2022.110008

Chien, C.-F., Hong Van Nguyen, T., Li, Y.-C., & Chen, Y.-J. (2023). Bayesian decision analysis for optimizing in-line metrology and defect inspection strategy for sustainable semiconductor manufacturing and an empirical study. *Computers & Industrial Engineering*, *182*, 109421. https://doi.org/10.1016/j.cie.2023.109421

Das, L., Gjorgiev, B., & Sansavini, G. (2024). Uncertainty-aware deep learning for monitoring and fault diagnosis from synthetic data. *Reliability Engineering & System Safety*, *251*, 110386. https://doi.org/10.1016/j.ress.2024.110386

Ding, Y., Jia, M., Cao, Y., Ding, P., Zhao, X., & Lee, C.-G. (2023). Domain generalization via adversarial out-domain augmentation for remaining useful life prediction of bearings under unseen conditions. *Knowledge-Based Systems*, *261*, 110199. https://doi.org/10.1016/j.knosys.2022.110199

Dong, M., Yang, A., Wang, Z., Li, D., Yang, J., & Zhao, R. (2024). Uncertainty-aware consistency learning for semi-supervised medical image segmentation. *Knowledge-Based Systems*, 112890. https://doi.org/10.1016/j.knosys.2024.112890

Fang, X., Easwaran, A., Genest, B., & Suganthan, P. N. (2024). *Your data is not perfect: Towards cross-domain out-of-distribution detection in class-imbalanced data*. https://arxiv.org/abs/2412.06284

Feng, M., Shao, H., Shao, M., Xiao, Y., Wang, J., & Liu, B. (2024). Utilizing Bayesian generalization network for reliable fault diagnosis of machinery with limited data. *Knowledge-Based Systems*, *305*, 112628. https://doi.org/10.1016/j.knosys.2024.112628

Germann, J. (2023). Global rivalries, corporate interests and Germany's 'National Industrial Strategy 2030'. *Review of International Political Economy*, *30*(5), 1749-1775. https://doi.org/10.1080/09692290.2022.2130958

Gital, Y., & Bilgen, B. (2024). Biomass supply chain network design under uncertainty, risk and resilience: A systematic literature review. *Computers & Industrial Engineering*, *193*, 110270. https://doi.org/10.1016/j.cie.2024.110270

Iaiani, M., Fazari, G., Tugnoli, A., & Cozzani, V. (2025). Identification of reference security scenarios from past event datasets by Bayesian Network analysis. *Reliability Engineering & System Safety*, *254*, 110615. https://doi.org/10.1016/j.ress.2024.110615

Jiang, D., He, C., Li, W., & Xu, D. (2025). Uncertainty-guided adversarial augmented domain networks for single domain generalization fault diagnosis. *Measurement*, *241*, 115674. https://doi.org/10.1016/j.measurement.2024.115674

Lee, Y., & Kang, S. (2024). Dynamic ensemble of regression neural networks based on predictive uncertainty. *Computers & Industrial Engineering*, *190*, 110011. https://doi.org/10.1016/j.cie.2024.110011

Li, H., Jiao, J., Liu, Z., Lin, J., Zhang, T., & Liu, H. (2025). Trustworthy Bayesian deep learning framework for uncertainty quantification and confidence calibration: Application in machinery fault diagnosis. *Reliability Engineering & System Safety*, *255*, 110657. https://doi.org/10.1016/j.ress.2024.110657

Liao, M., Shen, D., & Lv, P. (2023). A unified model of data uncertainty and data relation uncertainty. *Knowledge-Based Systems*, *278*, 110811. https://doi.org/10.1016/j.knosys.2023.110811

Lin, J., Shao, H., Min, Z., Luo, J., Xiao, Y., Yan, S., & Zhou, J. (2022). Cross-domain fault diagnosis of bearing using improved semi-supervised meta-learning towards interference of out-of-distribution samples. *Knowledge-Based Systems*, *252*, 109493. https://doi.org/10.1016/j.knosys.2022.109493

Lin, Y., Wang, Y., Zhang, M., & Zhao, M. (2025). A robust source-free unsupervised domain adaptation method based on uncertainty measure and adaptive calibration for rotating machinery fault diagnosis. *Reliability Engineering & System Safety*, *253*, 110516. https://doi.org/10.1016/j.ress.2024.110516

Liu, G., Liu, S., Li, X., Li, X., & Gong, D. (2025). Multiscenario deduction analysis for railway emergencies using knowledge metatheory and dynamic Bayesian networks. *Reliability Engineering & System Safety*, *255*, 110675. https://doi.org/10.1016/j.ress.2024.110675

Liu, G., & Wu, L. (2024). Incremental bearing fault diagnosis method under imbalanced sample conditions. *Computers & Industrial Engineering*, *192*, 110203. https://doi.org/10.1016/j.cie.2024.110203

Liu, W., & Yu, J. (2023). Densely connected semi-Bayesian network for machinery fault diagnosis with non-ideal data. *Mechanical Systems and Signal Processing*, *202*, 110678. https://doi.org/10.1016/j.ymssp.2023.110678

Liu, X., Chen, J., Zhang, K., Liu, S., He, S., & Zhou, Z. (2022). Cross-domain intelligent bearing fault diagnosis under class imbalanced samples via transfer residual network augmented with explicit weight self-assignment strategy based on meta data. *Knowledge-Based Systems*, *251*, 109272. https://doi.org/10.1016/j.knosys.2022.109272

Liu, Y., Ye, Z., Wang, R., Li, B., Sheng, Q. Z., & Yao, L. (2024). Uncertainty-aware pedestrian trajectory prediction via distributional diffusion. *Knowledge-Based Systems*, *296*, 111862. https://doi.org/10.1016/j.knosys.2024.111862

Mehimeh, S., Tang, X., & Zhao, W. (2023). Value function optimistic initialization with uncertainty and confidence awareness in lifelong reinforcement learning. *Knowledge-Based Systems*, *280*, 111036. https://doi.org/10.1016/j.knosys.2023.111036

Ren, Z., Zhu, Y., Kang, W., Fu, H., Niu, Q., Gao, D.,…Hong, J. (2022). Adaptive cost-sensitive learning: Improving the convergence of intelligent diagnosis models under imbalanced data. *Knowledge-Based Systems*, *241*, 108296. https://doi.org/10.1016/j.knosys.2022.108296

Snell, J., Swersky, K., & Zemel, R. (2017). Prototypical networks for few-shot learning. Proceedings of the 31st International Conference on Neural Information Processing Systems, Long Beach, California, USA.

Song, M., Moaveni, B., & Hines, E. (2025). Hierarchical Bayesian quantification of aerodynamic effects on an offshore wind turbine under varying environmental and operational conditions. *Mechanical Systems and Signal Processing*, *224*, 112174. https://doi.org/10.1016/j.ymssp.2024.112174

Sung, F., Yang, Y., Zhang, L., Xiang, T., Torr, P. H. S., & Hospedales, T. M. (2018). Learning to compare: Relation network for few-shot learning. 2018 IEEE/CVF Conference on Computer Vision and Pattern Recognition, Salt Lake City, UT, USA.

Tan, L., Huang, T., Liu, J., Li, Q., & Wu, X. (2023). Deep adversarial learning system for fault diagnosis in fused deposition modeling with imbalanced data. *Computers & Industrial Engineering*, *176*, 108887. https://doi.org/10.1016/j.cie.2022.108887

Vinyals, O., Blundell, C., Lillicrap, T., Kavukcuoglu, K., & Wierstra, D. (2016). Matching networks for one shot learning. Proceedings of the 30th International Conference on Neural Information Processing Systems, Barcelona, Spain.

Wang, R., Li, Y., Xu, J., Wang, Z., & Gao, J. (2022). F2G: A hybrid fault-function graphical model for reliability analysis of complex equipment with coupled faults. *Reliability Engineering & System Safety*, *226*, 108662.

https://doi.org/10.1016/j.ress.2022.108662
Wang, X., Li, L., Beck, J. L., & Xia, Y. (2021). Sparse Bayesian factor analysis for structural damage detection under unknown environmental conditions. *Mechanical Systems and Signal Processing*, *154*, 107563. https://doi.org/10.1016/j.ymssp.2020.107563
Wong, T.-T., & Tsai, H.-C. (2021). Multinomial naïve Bayesian classifier with generalized Dirichlet priors for high-dimensional imbalanced data. *Knowledge-Based Systems*, *228*, 107288. https://doi.org/10.1016/j.knosys.2021.107288
Xiao, D., Lin, Z., Yu, A., Tang, K., & Xiao, H. (2024). Data-driven method embedded physical knowledge for entire lifecycle degradation monitoring in aircraft engines. *Reliability Engineering & System Safety*, *247*, 110100. https://doi.org/10.1016/j.ress.2024.110100
Xu, L. D., Xu, E. L., & Li, L. (2018). Industry 4.0: State of the art and future trends. *International Journal of Production Research*, *56*(8), 2941-2962. https://doi.org/10.1080/00207543.2018.1444806
Xu, Q., Wang, Z., Jiang, C., & Jing, Z. (2024). Data-driven predictive maintenance framework considering the multi-source information fusion and uncertainty in remaining useful life prediction. *Knowledge-Based Systems*, *303*, 112408. https://doi.org/10.1016/j.knosys.2024.112408
Yin, J., Ren, X., Liu, R., Tang, T., & Su, S. (2022). Quantitative analysis for resilience-based urban rail systems: A hybrid knowledge-based and data-driven approach. *Reliability Engineering & System Safety*, *219*, 108183. https://doi.org/10.1016/j.ress.2021.108183
Yuan, Y., Wei, J., Huang, H., Jiao, W., Wang, J., & Chen, H. (2023). Review of resampling techniques for the treatment of imbalanced industrial data classification in equipment condition monitoring. *Engineering Applications of Artificial Intelligence*, *126*, 106911. https://doi.org/10.1016/j.engappai.2023.106911
Zhang, J., Zhang, K., An, Y., Luo, H., & Yin, S. (2023). An integrated multitasking intelligent bearing fault diagnosis scheme based on representation learning under imbalanced sample condition. *IEEE Transactions on Neural Networks and Learning Systems*, *35*(5), 6231-6242. https://doi.org/10.1109/TNNLS.2022.3232147
Zhang, T., Lin, J., Jiao, J., & Li, H. (2024). Cross-domain data fusion generation: A novel composite label-guided generative solution for adaptation diagnosis. *Knowledge-Based Systems*, *301*, 112284. https://doi.org/10.1016/j.knosys.2024.112284
Zhang, X., Meng, D., Gouk, H., & Hospedales, T. (2021). Shallow bayesian meta learning for real-world few-shot recognition. 2021 IEEE/CVF International Conference on Computer Vision (ICCV), Montreal, QC, Canada.

## Appendix 1: Task Implementation Process

| | **UBMF** |
|---|---|
| input | Meta-training set $D_{\text{meta - train}}$<br>Meta-test set $D_{\text{meta - test}}$<br>Task set $T_i$ : labeled training data $D^i_{\text{label}}$ unlabeled training data $D^i_{\text{unlabel,}}$, and test data $D^i_{\text{test}}$. |
| output | The current task diagnosis model $D_{\Phi^*}$ |
| Step 1 | Pseudo-Label Propagation Phase:<br>1. Initialize the feature network $f^p_\Phi$ and distance encoder $E_\Phi$. Set the meta-mini-batch size $b_{\text{meta}}$ and meta-training iterations $T_{\text{meta}}$.<br>2. Extract the diagnostic task $T_i$ from $D_{\text{meta - train,}}$ and calculate the classification prototype $P^i_c$ on $D^i_{\text{label}}$ using Equation (4.4).<br>3. Update the class prototype $P^i_c$ using Equation (4.19).<br>4. Compute the pseudo-label annotation loss $L^i_s(\theta,\Phi)$ on $D^i_{\text{test}}$ using Equations (4.20) and (4.21).<br>5. Update the parameters of $f^p_\Phi$ and $E_\Phi$ using the optimizer.<br>6. Repeat steps 2-5 for $T_{\text{meta}}$ iterations to complete the pseudo-label propagation phase. |
| Step 2 | Meta Self-Supervised Feature Extraction Phase<br>7. 1 Initialize the self-supervised feature extraction network $f^e_\Phi$.<br>8. Set the self-supervised training steps $T_{\text{sl,}}$, weak augmentation set $\{\delta_m\}$, strong augmentation set $\{\Delta_m\}$, and batch size $b_{\text{sl}}$.<br>9. Define temperature $\tau$, pseudo-label confidence threshold $\tau_p$, and loss balance coefficient $\lambda_w$.<br>10. From $D_{\text{meta - train,}}$ randomly sample class-balanced training data $\overleftarrow{D}_i$ for task $T_i$.<br>11. Apply weak augmentation $\delta_m$ to $X_k \in \overleftarrow{D}_i$, and extract feature vector $\xi_k$ and perturbation vector $\kappa_{k+B}$ |
| | using $f^e_\Phi$.<br>12. Compute the self-supervised loss $L^i_{ssl}$ .<br>13. Repeat steps 3-5 for $b_{meta}$ iterations, updating the network $f^e_\Phi$. |
| Step 3 | Task-Invariant Feature Extraction Phase<br>14. Extract features from $S_i$ (labeled data) and $\tilde{D}_i$.<br>15. Apply strong augmentation $\Delta_m$ to unlabeled samples and compute pseudo-supervised loss $L^i_c$ using Equation (4.24).<br>16. Update the self-supervised network $f^e_\Phi$.<br>17. Repeat steps 1-3 for $T_{sl}$ iterations. |
| Step 4 | Two-Level Classifier Bayesian Learning Phase<br>18. Initialize the knowledge vector $\phi$, where $\phi$:$\{\eta = 0, \psi = I, \lambda = 1, v = d\}$.<br>19. $d$ : dimension of the feature vector, $0$ : zero vector of dimension $d_t I$ :<br>20. $d \times d$ identity matrix.<br>Set mini-batch size $b_{QDA}$.<br>21. Extract task $T_i$ from $D_{meta\text{-}train}$ and target training set.<br>22. Extract the final feature vector $\xi_{final}$ using $f^{final}_\Phi$.<br>23. Update the task's classification parameters.<br>24. Repeat steps 2-4 until the training converges. |
| Step 5 | Diagnosis Model Output<br>25. Output the final diagnosis model:<br>$D_{\Phi^*} = QDA^{final}_{\Phi^*}\left(f^{final}_\Phi\left(X^i\right)\right)$ |
| | **Sample Filtering and Model Calibration Framework** |
| Input | Meta-training set $D_{meta\text{-}train}$<br>Meta-test set $D_{meta\text{-}test}$<br>Diagnostic tasks $T_i$ with:<br>Labeled training set $D^i_{label}$<br>Unlabeled training set $D^i_{unlabel}$<br>Test set $D^i_{test}$<br>Thresholds:<br>Distribution uncertainty threshold $\tau_{OOD}$<br>Confidence threshold $\tau_c$ |
| Output | Calibrated diagnosis model $D_{\Phi^*}$ |
| Step 1 | Distribution Uncertainty Quantification for OOD Detection<br>26. Initialize the feature network $f^p_\Phi$ and distance encoder $E_\Phi$.<br>27. From the meta-training set $D_{meta\text{-}train}$ extract diagnostic task $T_i$ and calculate class prototypes $P^i_c$ on $D^i_{label}$ using Equation (4.4).<br>28. Compute the uncertainty $U(X)$ for each sample $X_k \in D^i_{unlabel}$ using:<br>29. $U(X_k) = \text{Entropy}\left(p_\Phi(X_k)\right)$ or any uncertainty metric.<br>30. Filter OOD samples:<br>31. If $U(X_k) > \tau_{OOD}$, mark $X_k$ as OOD and remove it: $D^i_{filtered} = D^i_{unlabel} \setminus \{X_k \mid U(X_k) > \tau_{OOD}\}$. |
| Step 2 | :Confidence-Based Filtering of Low-Quality Samples<br>32. Extract filtered samples $D^i_{filtered}$ and compute the confidence score $p_\Phi(X_k)$ using the current network $f^p_\Phi$ :$p_\Phi(X_k) = \max_c P^i_c(X_k)$.<br>33. Filter low-confidence samples: If $p_\Phi(X_k) < \tau_c$, remove the sample $X_k$ : $D^i_{confident} = D^i_{filtered} \setminus \{X_k \mid p_\Phi(X_k) < \tau_c\}$. |
| Step 3 | Model Calibration Using Remaining Samples<br>34. Train the model: Use the confident samples $D^i_{confident}$ and labeled data $D^i_{label}$ to update the feature network $f^p_\Phi$ :<br>35. Minimize the pseudo-label loss $L^i_s(\theta,\Phi)$ (Equation 4.21).<br>36. Optimize the network parameters $\Phi$ using: $\Phi \leftarrow \Phi - \eta \nabla_\Phi L^i_s(\theta,\Phi)$. |
| Step 4 | Fine-Tuning with Class-Balanced Training<br>37. From $D_{meta\text{-}train}$, extract class-balanced training data $D^i_{balanced}$ and update the classification prototypes $P^i_c$ using Equation (4.19).<br>38. Perform weak augmentation $\delta_m$ and strong augmentation $\Delta_m$ on $D^i_{balanced}$.<br>39. Fine-tune the model using both labeled and confident samples. |
| Step 5 | Model Output<br>40. Perform final calibration on the test set $D^{i}_{test}$ to validate the model performance.<br>41. Output the calibrated diagnosis model: $D_{\Phi^*} = QDA^{final}_{\Phi^*}\left(f^{final}_{\Phi}\left(X^{i}\right)\right)$. |

## Appendix 2: Description of the Datasets

| Dataset Abbreviation | Operating Conditions | Fault Severity | Number of Health State Categories | Labeled Samples | Unlabeled Sample Distribution |
|---|---|---|---|---|---|
| UC_gearbox | Single Operating Condition | Multiple Severities | 5 | Healthy,<br>Missing,<br>Crack,<br>Spall,<br>Chip | 30,<br>10,<br>10,<br>10,<br>10 |
| DC_bearing | Variable Operating Conditions | Single Severity | 8 | Normal,<br>Roller,<br>Inner,<br>Outer,<br>Inner+Roller,<br>Outer+Inner,<br>Outer+Roller,<br>Outer+Inner+Roller | 40,<br>10,<br>10,<br>10,<br>10,<br>10,<br>10,<br>10 |
| IMS_bearing | Single Operating Condition | Multiple Severities | 4 | Normal，<br>IR，<br>OR，<br>RE | 100，<br>20，<br>30，<br>20 |
| CWRU_bearing | Multiple Operating Conditions | Multiple Severities | 4 | N,<br>OF,<br>IF,<br>RF | 100,<br>20,<br>20,<br>20 |
| PU_bearing | Multiple Operating Conditions | Multiple Severities | 4 | Healthy,<br>OR,<br>IR,<br>IR+OR | 100,<br>20,<br>20,<br>20 |
| UO_bearing | Multiple Operating Conditions | Single Severity | 3 | H,<br>I,<br>O | 100,<br>20,<br>20 |
| JNU_bearing | Multiple Operating Conditions | Single Severity | 4 | N,<br>IB,<br>OB,<br>TB | 100,<br>30,<br>20,<br>20 |
| MFPT_bearing | Multiple Operating Conditions | Single Severity | 3 | H,<br>OR,<br>IR | 100,<br>40,<br>20 |
| RIO_bearing | Multiple Operating Conditions | Single Severity | 4 | N,<br>IR,<br>OR,<br>RE | 100,<br>20,<br>20,<br>20 |
| XJTU_bearing | Multiple Operating Conditions | Single Severity | 4 | N,<br>IR,<br>OR, | 100,<br>20,<br>40, |

| Dataset Abbreviation | Operating Conditions | Fault Severity | Number of Health State Categories | Labeled Samples | Unlabeled Sample Distribution |
|---|---|---|---|---|---|
| | | | | Cage | 10 |